\documentclass{article} 
\pdfoutput=1
\usepackage{iclr2026_conference,times}

\usepackage[utf8]{inputenc}
\usepackage[T1]{fontenc}
\usepackage{graphicx}
\usepackage{booktabs}
\usepackage{amsmath,amssymb}
\usepackage{textcomp}
\usepackage{microtype}
\usepackage{xcolor}
\usepackage{tikz}
\usetikzlibrary{arrows.meta,positioning,fit,backgrounds}
\usepackage{hyperref}
\usepackage{url}

\iclrfinalcopy

\title{An Unexpected Robot Policy:\\
Early Evaluations of GPT-6 Astra\\
on RoboDojo and Beyond}

\author{%
Wenbo Zhang$^{1,2,*}$,\;
Kaixuan Wang$^{2,3,*}$,\;
Yutao Ouyang$^{1,4,*}$,\;
Xiaoyu Huang$^{1,5}$, \\
\bf Liyang Li$^{1}$,\;
Kailun Su$^{2,4}$,\;
Weiyang Jin$^{3}$,\;
Wenhao Chai$^{6}$, \\
\bf Haotian Liang$^{3}$,\;
Zhiyang Dou$^{7}$,\;
Yue Chen$^{2,8}$,\;
Tianxing Chen$^{2,3}$ \\
$^{1}$RoboProbe,\;
$^{2}$RoboDojo,\;
$^{3}$The University of Hong Kong, \\
$^{4}$Tsinghua University,\;
$^{5}$University of California, Berkeley, \\
$^{6}$Princeton University,\;
$^{7}$Massachusetts Institute of Technology,\;
$^{8}$Peking University \\
$^{*}$Equal contribution. \\
\makebox[\dimexpr\textwidth-2\tabcolsep\relax][c]{\small\url{https://robodojo-benchmark.com/report/gpt-6-astra-eval}}\\
\makebox[\dimexpr\textwidth-2\tabcolsep\relax][c]{\small\url{https://github.com/RoboProbe/RoboProbe}}%
}

\begin{document}

\maketitle
\lhead{Preprint}

\begin{abstract}
Embodied AI systems are often organized into System~1 and System~2.
System~1 is typically a pretrained policy that generates actions at high frequency,
whereas System~2 is often instantiated as a vision-enabled language model for
high-level planning. We ask whether a large language model (LLM) can act as the policy for
robot manipulation without task-specific finetuning. We call this setting LLM as
policy. We evaluate three LLMs on all 42 RoboDojo tasks and compare their scores
with 40 public policies. Astra and GPT-5.5 use the official 50-episode-per-task
protocol; DeepSeek-Flash uses 10 episodes per task. GPT-6 Astra achieves
22.48\% average success rate and 28.97 Score over 2,100 trials, ranking
above every public entry. Yet GPT-5.5 and DeepSeek-Flash reach only 0.88\% and 1.92\%
average success rate with the same post-processing. We find that Astra exhibits a sharply
polarized capability profile. It generalizes well to tasks that require semantic
understanding but not high-precision control. In contrast, it performs poorly on
tasks that require precision, dynamic control, or complex bimanual coordination.
In-context experiments show no aggregate benefit from one-shot demonstrations,
while selected interaction traces show within-episode corrections under perturbations.
Overall, the evaluated LLMs vary substantially in manipulation
performance. Astra stands out and provides initial evidence for the potential of a
general-purpose manipulation model, although reliable precision and dynamic
control remain limitations in the evaluated setting.
\end{abstract}

\section{Introduction}
\label{sec:introduction}

Robot manipulation systems are commonly organized as a hierarchy between fast
execution and slow deliberation~\citep{kahneman2011thinking}. System~1 is typically a pretrained policy that maps
observations and instructions to high-frequency actions~\citep{diffusionpolicy,act}. Vision Language Action
(VLA) models~\citep{rt2,openvla,pi0} and World Action Models
(WAMs)~\citep{dreamzero,cosmospolicy} instantiate this role.
System~2 is typically a vision-enabled language model that interprets the scene,
selects subgoals, or writes programs~\citep{palme,codeaspolicies}. Practical systems combine the two: the language
model decides what should happen, and a pretrained policy or scripted skill executes
the decision. SayCan, Inner Monologue, Code as Policies, and VoxPoser follow this
hierarchical pattern~\citep{saycan,innermonologue,codeaspolicies,voxposer}.

This division has been challenged repeatedly. An early example used
GPT-4~\citep{gpt4} to generate dense end-effector trajectories without motion primitives or robot
fine-tuning, but it relied on separate detection and segmentation models and a
30-task evaluation~\citep{kwon2024trajectory}. More recent reports suggest that
direct control is improving. Anthropic found that frontier models made increasing
subgoal progress on LIBERO-40, although full-task success remained between 0 and
5.5\%~\citep{liu2023libero,berman2026robotics}. Reports released with GPT-6
Astra~\citep{gpt6astra} showed strong
results on selected real-robot tasks, but used small custom task suites and different
interfaces~\citep{robocurveastra2026,agentaspolicy2026}. These studies establish
feasibility and motivate a sharper question, but they do not place frontier LLMs
against a complete public policy leaderboard under one benchmark protocol.

\begin{figure}[t]
  \centering
  \includegraphics[width=0.862\linewidth]{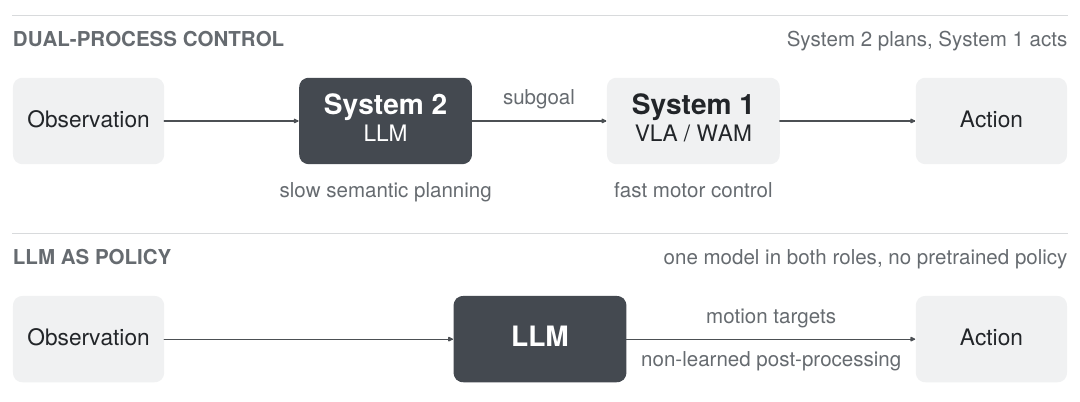}
  \caption{Two ways to turn an observation into an executable action.
  Top: a dual-process controller separates a language model for semantic planning
  from a pretrained motor policy for execution. VLA denotes a vision language
  action model; WAM denotes a world action model.
  Bottom: in the LLM-as-policy setting, the language model selects motion targets
  directly. Non-learned post-processing converts these targets into executable
  commands without a pretrained motor policy.}
  \label{fig:systems}
\end{figure}

We ask whether a frontier large language model (LLM) can itself serve as the policy for manipulation. We
call this setting \emph{LLM as policy}: every executable motion target is selected by
the model, with only simple non-learned post-processing before execution. Real-robot
tests alone cannot support the required comparison. They are slow, hard to scale,
and unsafe actions can stop data collection; in our tests, such actions damaged
equipment (Section~\ref{sec:real}). Deploying dozens of pretrained baselines on the
same hardware is also impractical. We therefore use RoboDojo simulation for the
primary evaluation and hardware for diagnostic evidence. RoboDojo provides 42 tasks,
five capability axes, and a fixed 50-episode protocol, yielding 2,100 trials for each
fully evaluated model and a public board of 40 pretrained
policies~\citep{robodojo2026}. Figure~\ref{fig:systems}
contrasts the two system designs; Appendix~\ref{sec:limitations} states the remaining
input and serving differences.

The answer is model-specific. GPT-6 Astra reaches 22.48\% average success rate (SR)
and 28.97 Score over 2,100 trials, ranking above all 40 public policies. This
confirms at benchmark scale the manipulation ability suggested by recent Astra
reports. Yet GPT-5.5 and DeepSeek-Flash reach only 0.88\% and 1.92\% average SR with
the same action interface and post-processing. This approximately 25-fold spread
shows that the shared action conversion alone does not explain the performance
difference. The positive results should not be generalized to all frontier LLMs.
To our knowledge, this
is the first complete public-benchmark evaluation to rank an LLM-as-policy controller
against the benchmark's full public policy board.

We therefore analyze Astra as a case study. Its aggregate lead hides a sharply split
capability profile: it is strongest on semantic and open-ended tasks, but remains
weak on contact-rich, precision-critical, dynamic, and coordinated bimanual tasks
(Section~\ref{sec:results-sim}). In our one-shot experiments, demonstrations reduce
aggregate success under the tested protocol. Separately, selected perturbation
traces show Astra revising its actions after observing their outcomes
(Section~\ref{sec:icl}). These observations are consistent with within-episode
adaptation, but do not isolate its mechanism or establish an advantage over
pretrained policies. Astra thus challenges a strict System~1/System~2 division without
eliminating the need for fast and precise pretrained control.

\section{Related Work}
\label{sec:related}

\paragraph{Large pretrained robot policies.} Vision language action (VLA) models
map visual observations and language instructions to robot
actions~\citep{rt1,openx,octo}. RT-2 established
web-scale vision--language transfer to control, OpenVLA provided an open 7B model, and
$\pi_0$ paired a vision--language backbone with a flow-matching action
expert~\citep{rt2,openvla,pi0}. World action models (WAMs) additionally predict future
visual states: DreamZero jointly generates video and actions, while Cosmos Policy
represents future observations, actions, and values in a shared video-model latent
space~\citep{dreamzero,cosmospolicy}.

\paragraph{Language models as high-level planners.} A second line uses a language
model to guide a specialized controller. SayCan selects learned skills, Inner
Monologue plans with language, Code as
Policies writes programs over control APIs, and VoxPoser constructs spatial value
maps for a motion planner~\citep{saycan,innermonologue,codeaspolicies,voxposer}. Hi Robot makes this
hierarchy explicit by generating language subgoals for a low-level VLA
policy~\citep{hirobot}. Helix instead passes a continuous semantic latent from a slow
vision--language model to a 200\,Hz visuomotor policy~\citep{helix}. In both designs,
the slower model guides execution but does not issue the executable targets.

\paragraph{Language models as policies.} Direct action generation predates the
current generation of frontier models. GPT-4 generated dense end-effector
trajectories without motion primitives or robot fine-tuning, but relied on separate
detection and segmentation models~\citep{kwon2024trajectory}. Recent evaluations show
that frontier models can also issue actions directly from
robot observations, while reliability remains uneven across
tasks~\citep{berman2026robotics,projectfetch2026,robocurveastra2026,agentaspolicy2026,faea2026}. These studies establish
feasibility, not a general replacement for pretrained control.

\paragraph{In-context robot learning.} Prior work commonly treats a demonstration
as the context from which a robot should infer a task. ICRT uses sensorimotor
trajectories, RoboPrompt uses textual action examples, and HOST, Skild S1, and
GEN-1.5 use one-shot human or sensorimotor demonstrations~\citep{fu2024icrt,
yin2024roboprompt,host,skilds1,gen15}. RoboTTT extends this paradigm with fast
weights that absorb long demonstration or interaction histories~\citep{robottt}.
A related line adapts from deployment interaction by inferring latent system
configurations, conditioning on histories across trials, or optimizing latent prompts
from interaction data~\citep{icwm,locoformer,tttvla}. Our perturbation experiments
examine whether GPT-6 Astra revises actions in response to feedback within an
episode, without updating its weights or carrying memory across episodes.

\section{Study design: evaluating LLM as policy}
\label{sec:study-design}

\subsection{What counts as LLM as policy}
\label{sec:study-boundary}

We define the LLM-as-policy setting using two rules. First, no learned policy
appears in the control path between the language model and the robot. Second, the
language model's weights remain fixed, with no robot-specific or task-specific
fine-tuning performed for this evaluation. The main benchmark runs are zero-shot;
Section~\ref{sec:icl} separately studies demonstrations and interaction history.
A model fine-tuned to map robot
observations to actions would instead fall under the conventional definition of a
vision language action model. The model may act through either Cartesian
end-effector targets or joint-space targets. For end-effector control, an inverse
kinematics solver may convert the target into executable joint
commands; this conversion is geometric only and does not involve a learned policy.

\subsection{Our implementation}
\label{sec:study-postprocessing}

\begin{figure}[t]
  \centering
  \includegraphics[width=\linewidth]{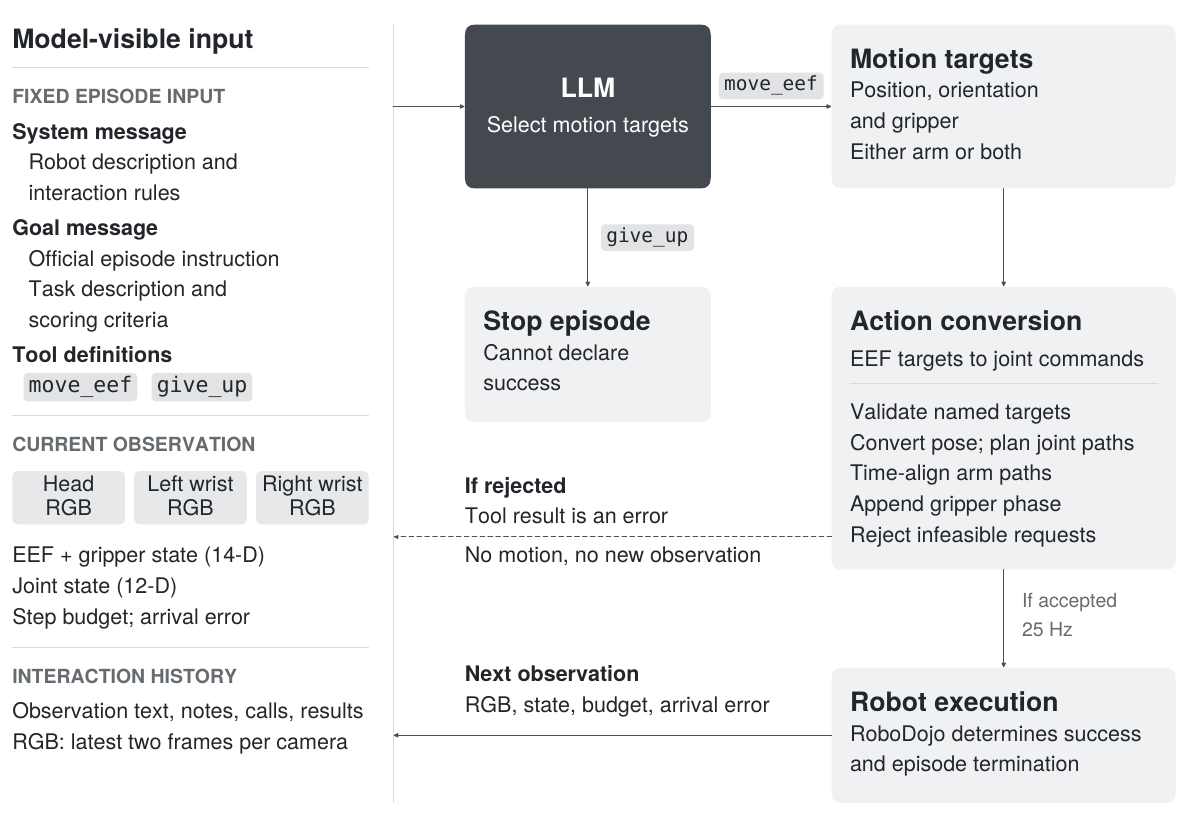}
  \caption{The model-facing context and execution loop.
  A frozen language model chooses world-frame end-effector (EEF) targets at
  the grasp point through \texttt{move\_eef}. The action conversion module is
  non-learned. It validates the request, converts grasp-point targets to flange
  poses, plans joint paths, and constructs executable joint commands.
  Requested gripper changes follow the arm-motion phase.
  The two return routes are the two branches of a request. A rejected target
  returns an error as the tool result and causes no motion, so the model
  retries against the same observation; this is the dashed route. An accepted
  target returns its planned duration, executes, and only then produces the
  next observation with the achieved state and arrival error; this is the solid
  route. \texttt{give\_up} requests termination, but only the environment
  determines success. The context retains the latest two RGB frames from each
  camera.}
  \label{fig:model-input}
\end{figure}

\noindent\textbf{Prompt composition.}
Figure~\ref{fig:model-input} summarizes our closed-loop implementation. Each episode
starts with a fixed system message, Goal message, and tool definitions. The system
message specifies the controller role, interaction rules, and embodiment. The Goal
message combines the official task instruction with a \emph{task recipe} derived
from the RoboDojo wiki. The recipe describes the task and its scoring stages without
providing an action sequence. The system message also provides task-independent
operating advice, including using an idle wrist camera to inspect the work area.
\newline\textbf{Observation and context management.}
At each turn, the model receives the dynamic text history and the current observation.
The text history retains all prior observation text, model-written notes, tool calls, and tool
results. Each observation contains three RGB views, the remaining step budget, and
the robot state. The state consists of 14 grasp-point dimensions and 12 read-only
joint angles. RGB images are retained only for the two most recent observation turns.
\newline\textbf{Tools.}
The model must call either \texttt{move\_eef} or \texttt{give\_up} at every turn.
\texttt{move\_eef} requests a robot motion, while \texttt{give\_up} ends an episode
that the model judges unrecoverable. We do not provide pick, place, or pregrasp
primitives because they would solve part of the task outside the model.
\newline\textbf{Action space.}
The \texttt{move\_eef} target may specify any subset of the position, orientation,
and gripper dimensions for either arm. Unspecified dimensions retain their observed
values. Specifying both arms produces a simultaneous motion. Appendix~\ref{sec:app-prompt}
summarizes the tools, units, and bounds.
\newline\textbf{Action conversion and feedback.}
Non-learned software rejects invalid targets and attempts to plan a joint-space
path to each valid target. Accepted paths are resampled for execution at 25\,Hz.
The interface reports whether the target was accepted and the planned execution
duration. RoboDojo executes the path and returns the next observation, including
the discrepancy between the requested target and the measured end-effector state.
This conversion is not a guarantee of collision-free or safe execution.
Appendix~\ref{sec:app-prompt} describes the model-facing feedback.
\newline\textbf{Information boundary and termination.}
The model receives no metric depth, privileged object pose, reward state, layout
specification, or layout-specific solution. \texttt{give\_up} does not declare
success or change the reward. Success is determined only by the RoboDojo environment.

\subsection{Evaluation protocol}
\label{sec:study-protocol}

The benchmark LLM runs use the same action space, post-processing, observation
format, and context policy, with one evaluation seed per model.
We evaluate GPT-6 Astra and GPT-5.5 for 50
episodes on each of the 42 tasks, following the official
protocol~\citep{robodojo2026}. This gives 2,100
trials per model. Because of evaluation cost, we evaluate DeepSeek-Flash for 10
episodes per task, giving 420 trials in total. 
For each Generalization task, it runs five standard and five randomized
episodes. We mark all of its results with
\dag{}. All benchmark LLM runs use medium reasoning effort and a default budget of 100 model calls per episode.
We compare against the public large pretrained robot policies as recorded on the leaderboard on 
September 10, 2026~\citep{community2026xpolicylab}.

Score is mean process reward multiplied by 100; SR is the percentage of episodes
that satisfy the environment's full-success criterion. The leaderboard Average
is the unweighted mean of the five axis-level values. It differs from pooling
all episodes because the axes contain different numbers of tasks. Generalization
cells average the standard and randomized conditions. The public policies were
not rerun with the wiki-derived text supplied to the LLMs, so the comparison
shares outcome metrics but not identical input information.

\section{How far can LLM as policy go?}
\label{sec:results-sim}

\subsection{GPT-6 Astra Leads Overall, While Other LLMs Lag Behind}
\label{sec:results-board}

Table~\ref{tab:board-compact} ranks the three LLM controllers alongside the 40 public
policies in RoboDojo's official Score/SR\% cell format. GPT-6 Astra takes
rank 1 at 28.97 Score / 22.48\% Average SR (micro $472/2100 = 22.48\%$ SR, 28.72
Score), ahead of all 40 public policy entries in this comparison.
GPT-5.5 reaches 1.13/0.88\% and DeepSeek-Flash 2.99\dag/1.92\%\dag.
The highest and lowest Average SR among the three LLMs differ by approximately
25-fold.

\begin{table}[htbp]
  \centering
  \caption{RoboDojo-Sim board, Score/SR\% per cell: the top ten of 43 ranked
  entries plus the two remaining LLM controllers, which rank 28 and 33. Ranks
  are over all 43; the public submission contains only the 40 policy rows, where the
  same order gives DM0.5 rank~1. \dag~DeepSeek-Flash is 1 seed $\times$ 10 episodes
  per task, not 50. Policy rows are a 2026-09-10 leaderboard snapshot. The full
  43-row board is Table~\ref{tab:sim-board}.}
  \label{tab:board-compact}
  \small
  \setlength{\tabcolsep}{4pt}
  \resizebox{\linewidth}{!}{%
  \begin{tabular}{rlcccccc}
    \toprule
    Rank & Model & Average & Gen. & Prec. & Long-Hor. & Memory & Open \\
    \midrule
    1 & \textbf{GPT-6 Astra} & 28.97/22.48 & 33.36/30.50 & 12.65/4.00 & 21.45/8.25 & 43.04/38.67 & 34.36/31.00 \\
    2 & DM0.5 & 24.90/19.34 & 15.77/10.95 & 24.82/16.75 & 33.70/19.50 & 47.74/47.44 & 2.43/2.08 \\
    3 & GalaxeaVLA (G0.5) & 20.23/14.88 & 18.46/12.83 & 28.25/20.42 & 44.12/32.25 & 8.61/7.33 & 1.73/1.58 \\
    4 & Xiaomi-Robotics-1 & 20.07/13.93 & 23.54/17.00 & 26.69/18.83 & 38.39/23.67 & 7.81/6.56 & 3.94/3.58 \\
    5 & OpenWAM-$\alpha$ & 17.18/11.92 & 20.71/14.83 & 18.45/9.25 & 34.93/25.33 & 10.41/9.11 & 1.41/1.08 \\
    6 & Meituan-Robotics-0 & 14.95/9.53 & 13.75/8.17 & 16.77/7.75 & 29.61/18.58 & 10.06/8.89 & 4.54/4.25 \\
    7 & Hy-Embodied-0.5-VLA & 13.07/8.80 & 11.78/8.39 & 13.81/8.00 & 25.74/14.92 & 13.37/12.11 & 0.65/0.58 \\
    8 & Spatial Forcing & 12.38/8.04 & 14.12/9.34 & 17.32/10.58 & 23.26/14.58 & 5.43/4.11 & 1.78/1.58 \\
    9 & Pi-05 & 11.41/6.91 & 13.38/8.17 & 12.40/5.50 & 23.54/14.67 & 5.78/4.56 & 1.98/1.67 \\
    10 & InternVLA-A1.5 & 11.15/7.14 & 10.35/6.83 & 15.23/10.17 & 23.80/13.75 & 4.93/3.56 & 1.43/1.42 \\
    \addlinespace
    28 & \textbf{DeepSeek-Flash\dag} & 2.99/1.92 & 2.42/1.67 & 1.75/0.00 & 2.12/0.00 & 2.17/1.67 & 6.50/6.25 \\
    33 & \textbf{GPT-5.5} & 1.13/0.88 & 0.19/0.00 & 0.42/0.00 & 0.51/0.00 & 1.67/1.67 & 2.87/2.75 \\
    \bottomrule
  \end{tabular}}
\end{table}

The remaining analyses focus on Astra as a case study rather than on LLMs in general.

\subsection{Zero-shot Evaluation}
\label{sec:results-axis}

\subsubsection{A Heavily Imbalanced Policy}
Astra achieves an Average SR of 22.48\%, exceeding DM0.5, the highest-ranked
public policy, by 3.14 percentage points. This lead is driven by Open and Generalization.
In comparison, it performs worse than DM0.5 on Memory, Precision, and Long-Horizon tasks.
Table~\ref{tab:f2} shows complementary task-level outcomes, including tasks on
which Astra records no success despite high public-policy success rates.
Its overall lead therefore reflects strengths
in particular tasks alongside substantial gaps in others, rather than
consistently reliable performance.
\begin{table}[htbp]
  \centering
  \caption{Complementary task-level outcomes. Left: Astra reaches at least 20\%
  SR while every public policy remains below 5\%. Right: the strongest public
  policy reaches at least 20\% while Astra remains below 5\%. Xiaomi denotes
  Xiaomi-Robotics-1; G0.5, GalaxeaVLA (G0.5).}
  \label{tab:f2}
  \scriptsize
  \setlength{\tabcolsep}{2.5pt}
  \begin{tabular}{@{}lrr@{\qquad}lrr@{}}
    \toprule
    \multicolumn{3}{c}{\textit{Astra $\ge$ 20\%, policy $<$ 5\%}} &
    \multicolumn{3}{c}{\textit{Policy $\ge$ 20\%, Astra $<$ 5\%}} \\
    Task & Astra & Best policy & Task & Astra & Best policy \\
    \midrule
    \texttt{push\_T}                       & 60.0 & 0.7 Xiaomi &
    \texttt{make\_kong}                    & 0.0 & 90.0 G0.5 \\
    \texttt{arrange\_largest\_number}      & 60.0 & 4.7 Xiaomi &
    \texttt{build\_tower}                  & 2.0 & 78.7 G0.5 \\
    \texttt{align\_blocks}                 & 50.0 & 0.0 DM0.5 &
    \texttt{insert\_tubes}                 & 0.0 & 59.3 DM0.5 \\
    \texttt{solve\_equation}               & 40.0 & 0.0 DM0.5 &
    \texttt{play\_tic\_tac\_toe}           & 0.0 & 58.7 OpenWAM \\
    \texttt{stack\_blocks\_by\_language}   & 40.0 & 2.7 DM0.5 &
    \texttt{pour\_balls\_into\_vase}       & 4.0 & 46.0 Xiaomi \\
    \texttt{imitate\_sorting\_sequence}    & 36.0 & 0.7 Xiaomi &
    \texttt{pour\_liquid\_into\_cup}       & 4.0 & 42.0 Xiaomi \\
    \texttt{classify\_objects\_by\_language} & 30.0 & 0.7 Xiaomi &
    \texttt{store\_laptop\_and\_headphones} & 0.0 & 22.7 Xiaomi \\
    & & & \texttt{fill\_pen\_holder}       & 0.0 & 22.0 Xiaomi \\
    \bottomrule
  \end{tabular}
\end{table}

\subsection{Qualitative zero-shot capabilities}
Astra exhibits several behaviours without a pretrained motor policy or scripted
manipulation skill. Figure~\ref{fig:capability}(a--c)
illustrates these through three qualitative episodes:

\noindent\textbf{Semantic understanding.}
In \texttt{make\_kong}, Astra selects the tile from its own collection that
matches the opponent's play.\\
\textbf{Active perception.}
In \texttt{insert\_tubes}, Astra uses active perception~\citep{bajcsy1988active} by repositioning
the opposite wrist camera to view the rack when the held tube occludes
the other camera. This behaviour follows the general active-view advice in the
system prompt; it is not evidence of discovering that strategy without guidance.\\
\textbf{Flexible use of both arms.}
In \texttt{classify\_objects\_by\_language}, Astra holds a different object
in each hand, demonstrating concurrent use of both arms~\citep{act}.

\noindent These examples establish that the behaviors occur, but do not
quantify their frequency or contribution to task success.

\subsection{Physical boundary.}
\label{sec:results-wall}
Observed failures include errors in lift height, approach angle, or correction
timing, even in episodes where the model appears to interpret the task correctly.
Following~\citet{physicalcommonsense}, we characterize these difficulties as
limitations in \emph{physical commonsense}~\citep{battaglia2013simulation}: an implicit grasp of contact
dynamics, force sensitivity, and collision geometry. These weaknesses qualify
the interpretation of its aggregate rank, which alone does not establish that
LLM-based policies can replace pretrained motor policies. The observations do not
separate missing physical knowledge from limitations of perception or the action
interface (Appendix~\ref{sec:limitations}).
Figure~\ref{fig:capability}(d--f) illustrates three such limitations through
qualitative failure episodes:

\noindent\textbf{Dynamic control.}
In \texttt{pick\_from\_conveyor\_by\_image}, Astra misses a moving target
during grasping, illustrating the challenge of coordinating motion
with a changing scene.\\
\textbf{Bimanual coordination.}
In \texttt{sweep\_blocks}, a failed two-hand transfer shows that engaging
both arms does not ensure coordinated manipulation.\\
\textbf{Awareness of surrounding objects.}
In \texttt{build\_tower}, Astra knocks over the existing structure while
reaching for the next block, illustrating a failure to account for
surrounding objects along its motion path.

\noindent These episodes identify failure modes that aggregate success
rates alone do not distinguish.

\makeatletter
\@ifundefined{capfigwidth}{\newlength{\capfigwidth}}{}
\makeatother
\begin{figure}[t]
  \centering
  \setlength{\capfigwidth}{0.49\linewidth}
  \newcommand{\capcolhead}[1]{%
    {\small\bfseries\scshape #1}\\[0.3mm]
    \rule{\linewidth}{0.4pt}\\[0.6mm]}
  \newcommand{\cappanel}[4]{%
    \includegraphics[width=\linewidth]{#1}\\[-0.4mm]
    \begin{minipage}[t]{0.333\linewidth}
      \centering{\scriptsize Left wrist}
    \end{minipage}%
    \begin{minipage}[t]{0.333\linewidth}
      \centering{\scriptsize Head}
    \end{minipage}%
    \begin{minipage}[t]{0.333\linewidth}
      \centering{\scriptsize Right wrist}
    \end{minipage}\\[-0.2mm]
    {\scriptsize\bfseries #2}\\[-0.2mm]
    {\scriptsize\textcolor{black!65}{\texttt{#3}}}\\[-0.2mm]
    {\scriptsize #4}}

  \begin{minipage}[t]{\capfigwidth}
    \centering
    \capcolhead{Capabilities}
    \cappanel{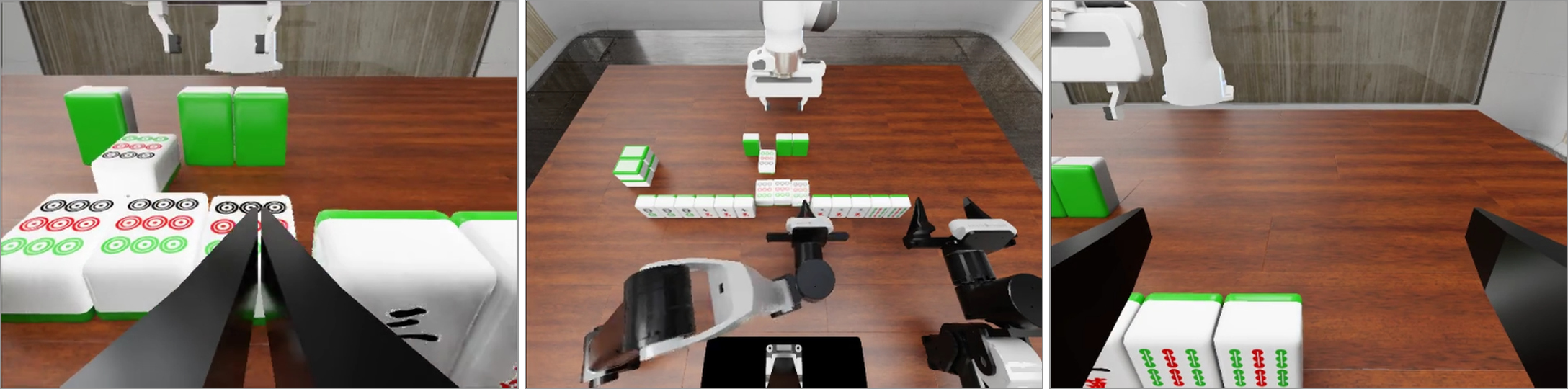}
      {(a) Semantic matching}
      {make\_kong}
      {opponent's tile matched}\\[3mm]
    \cappanel{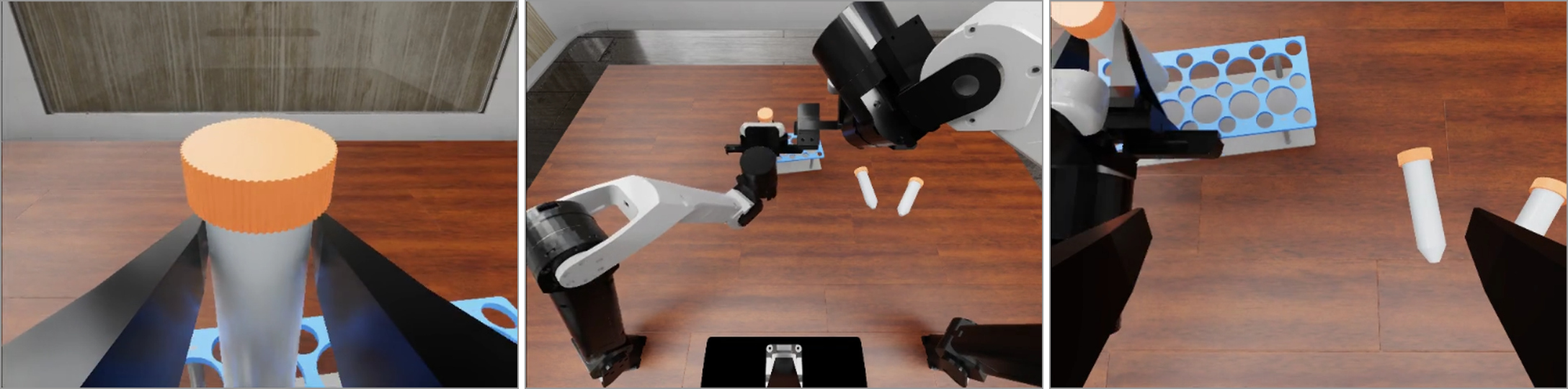}
      {(b) Active vision}
      {insert\_tubes}
      {occluded rack view recovered}\\[3mm]
    \cappanel{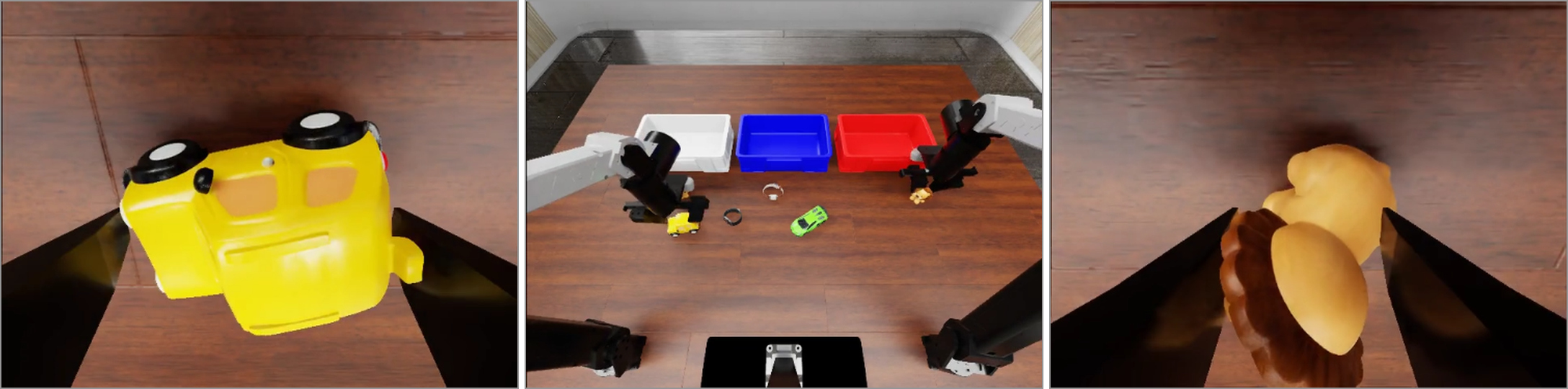}
      {(c) Flexible bimanual roles}
      {classify\_objects\_by\_language}
      {one object carried by each hand}
  \end{minipage}%
  \hfill
  \begin{minipage}[t]{\capfigwidth}
    \centering
    \capcolhead{Limits}
    \cappanel{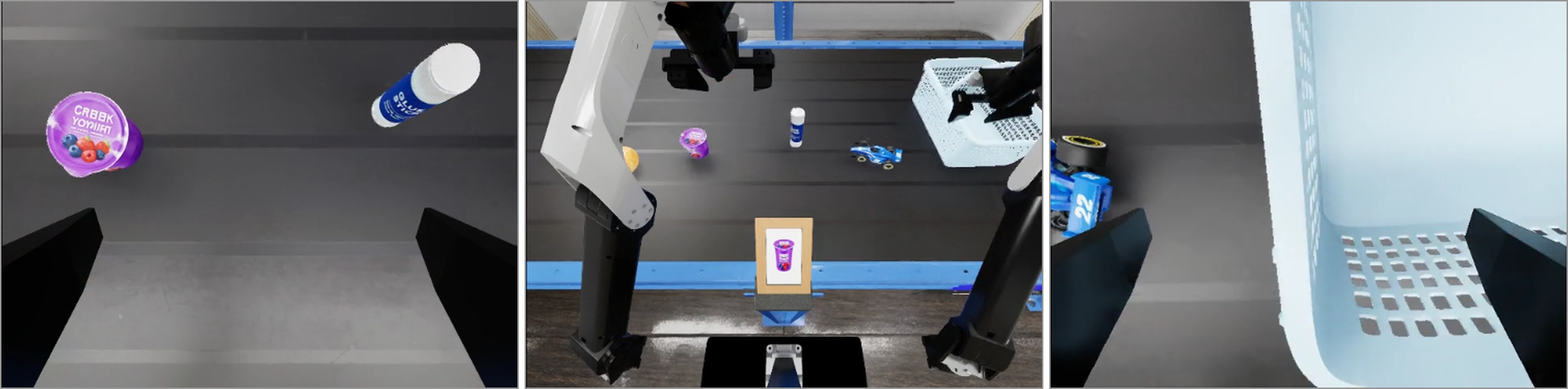}
      {(d) Precision and dynamic control}
      {pick\_from\_conveyor\_by\_image}
      {moving target missed}\\[3mm]
    \cappanel{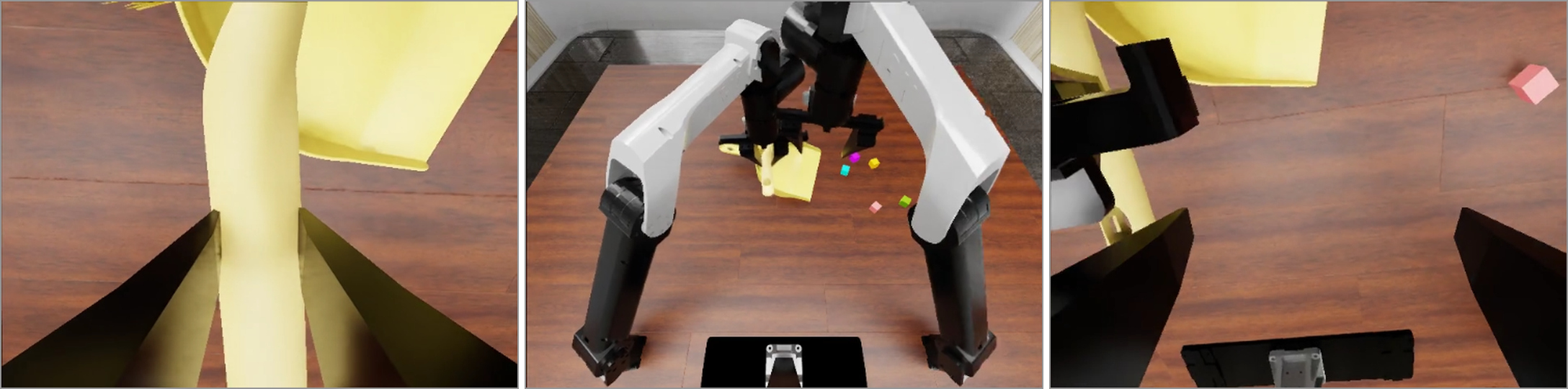}
      {(e) Bimanual coordination}
      {sweep\_blocks}
      {two-hand transfer fails}\\[3mm]
    \cappanel{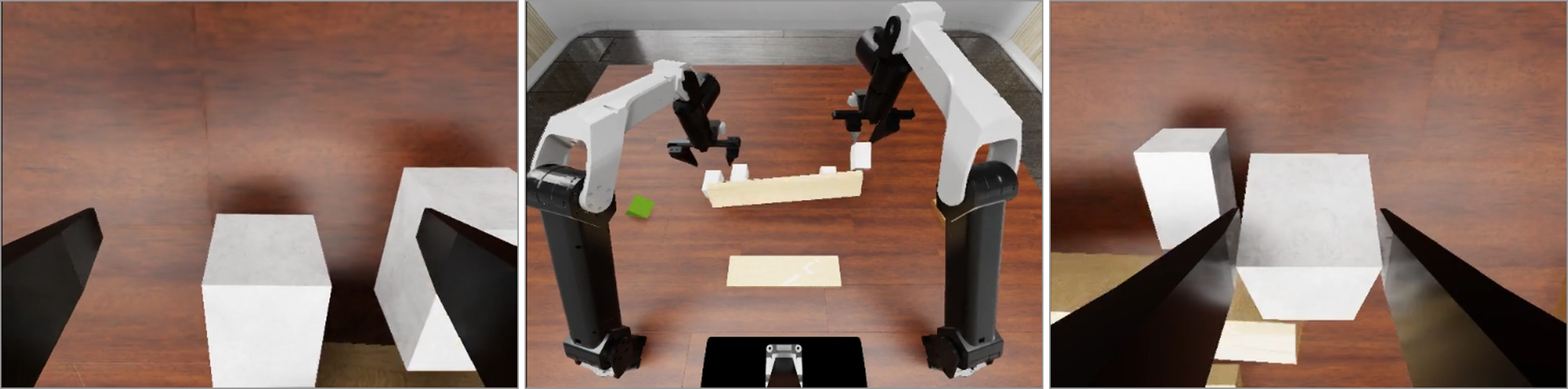}
      {(f) Surrounding-object awareness}
      {build\_tower}
      {built structure struck while reaching}
  \end{minipage}

  \caption{Capabilities and limits of Astra used as a policy. Each panel shows one
  episode in three camera views; the task and observed evidence appear below.
  Panels (a)--(c) show capabilities and panels (d)--(f) show limits. These examples
  establish occurrence, not frequency.}
  \label{fig:capability}
\end{figure}

\subsection{In-context learning from demonstrations and interaction}
\label{sec:icl}

We examine in-context learning (ICL)~\citep{brown2020language,min2022rethinking}
through demonstrations and interaction histories, two forms of context studied in
robot learning~\citep{fu2024icrt,yin2024roboprompt,host,skilds1,gen15,robottt,icwm,locoformer,tttvla}.
One-shot demonstrations do not improve Astra's aggregate performance in our probe.
Selected interaction traces instead show corrections within an episode.
These experiments remain a case study of Astra, not a claim about LLMs in general.

\phantomsection
\noindent\textbf{One-shot demonstrations do not improve aggregate performance here.}
\label{sec:icl-demo}
We first test demonstration-conditioned ICL in a one-shot setting. The model
receives one example of the desired behaviour. We evaluate 340 matched task--layout
pairs across 34 tasks. The example comes from a different layout of the same task.
One condition provides images and corresponding end-effector poses. The other
describes the same trajectory in text. The zero-shot baseline uses the same 340 pairs
from the official campaign. Appendix~\ref{sec:app-oneshot} gives protocol details and
task-level evidence.

\begin{table}[htbp]
  \centering
  \caption{Demonstration ICL on 340 matched task--layout pairs: the same 34 tasks
  and ten layouts per task under each condition. The two demonstration conditions
  are new runs of 340 episodes each; the zero-shot baseline is the official
  50-episode Astra campaign restricted to those exact pairs.}
  \label{tab:f3}
  \small
  \begin{tabular}{lrlr}
    \toprule
    Condition & Episodes & SR & vs zero-shot \\
    \midrule
    Zero-shot (baseline) & 340 & 78/340 = 22.9\% & --- \\
    Image + end-effector demonstration & 340 & 61/340 = 17.9\% & $-$5.0 \\
    Text demonstration & 340 & 44/340 = 12.9\% & $-$10.0 \\
    \bottomrule
  \end{tabular}
\end{table}

Neither demonstration condition improves success over the matched 22.9\% zero-shot
baseline (Table~\ref{tab:f3}). The task-level split shows that demonstrations rarely
produce successes on tasks with no success in the matched zero-shot runs: the
image condition succeeds in 2 of these 150 episodes, and the text condition in none. The
aggregate decline instead comes from losing episodes on tasks with zero-shot
successes (Appendix~\ref{sec:app-oneshot-subsets}).

One possible explanation is that a demonstration adds little when execution
precision, rather than task interpretation, limits performance. Another is that
the model transfers layout-specific geometry from the example to the current
scene. These are hypotheses: the experiment does not isolate either cause from
effects of demonstration format, context length, or stochastic variation.

\phantomsection
\noindent\textbf{Within-episode correction under perturbations.}
\label{sec:icl-online}
We next probe whether Astra revises actions in response to interaction feedback,
without a demonstration. We apply six perturbations to
eight \texttt{general\_pickup} layouts. The perturbations alter
visual observations, action execution, or the mapping of Cartesian coordinates to
executed motion. The model weights and context policy remain fixed. No explicit
explanation of the perturbation is added, but the negated-coordinate condition
changes the bounds shown in the tool description. Appendix~\ref{sec:app-perturbation}
defines this information boundary and the protocol.

\begin{table}[htbp]
  \centering
  \caption{Performance under perturbations on eight \texttt{general\_pickup} layouts.
  Visual perturbations transform or remove camera views. Per-move pose jitter offsets
  each executed pose by a random displacement with mean magnitude 10\,cm. Negated
  Cartesian axes change the coordinate convention and the advertised target bounds. The layouts are
  the subset Astra solves without perturbation, and each cell contains one
  episode per layout.}
  \label{tab:f4}
  \small
  \begin{tabular}{llc}
    \toprule
    Perturbed component & Condition & Successes \\
    \midrule
    \multicolumn{3}{l}{\textit{LLM as policy (GPT-6 Astra)}} \\
    None               & No perturbation             & 8/8 \\
    Visual observation & Top--bottom image flip       & 8/8 \\
    Visual observation & Left--right image mirror     & 6/8 \\
    Visual observation & No head-camera view          & 6/8 \\
    Visual observation & Right-wrist view only        & 3/8 \\
    Action execution   & Per-move pose jitter          & 3/8 \\
    Coordinate mapping & Negated Cartesian axes        & 4/8 \\
    \bottomrule
  \end{tabular}
\end{table}

Table~\ref{tab:f4} reports task success under these perturbations. Astra retains
some success in every condition. Success alone does not distinguish existing
robustness, repeated attempts, and adaptation from feedback. We therefore examine
selected traces for changes in the model's actions and stated interpretation.
The small probe does not establish the relative difficulty of the conditions.

\makeatletter
\@ifundefined{pvcellwidth}{\newlength{\pvcellwidth}}{}
\@ifundefined{pvcellheight}{\newlength{\pvcellheight}}{}
\makeatother
\begin{figure}[t]
  \centering
  \setlength{\pvcellwidth}{\dimexpr(\linewidth-22mm)/6\relax}
  \setlength{\pvcellheight}{0.75\pvcellwidth}
  \newcommand{\pvcell}[1]{%
    \fcolorbox{black!55}{black!4}{%
      \parbox[c][\pvcellheight][c]{\pvcellwidth}{\centering\scriptsize
        \textcolor{black!45}{#1}}}}
  \newcommand{\pvimg}[1]{%
    \parbox[c][\dimexpr\pvcellheight+2\fboxsep+2\fboxrule\relax][c]{%
        \dimexpr\pvcellwidth+2\fboxsep+2\fboxrule\relax}{%
      \includegraphics[width=\dimexpr\pvcellwidth+2\fboxsep+2\fboxrule\relax,
        height=\dimexpr0.75\pvcellwidth+1.5\fboxsep+1.5\fboxrule\relax]{figures/assets/#1}}}
  \newcommand{\pvimgann}[2]{%
    \parbox[c][\dimexpr\pvcellheight+2\fboxsep+2\fboxrule\relax][c]{%
        \dimexpr\pvcellwidth+2\fboxsep+2\fboxrule\relax}{%
      \begin{tikzpicture}
        \node[anchor=south west,inner sep=0] (img) at (0,0) {%
          \includegraphics[width=\dimexpr\pvcellwidth+2\fboxsep+2\fboxrule\relax,
            height=\dimexpr0.75\pvcellwidth+1.5\fboxsep+1.5\fboxrule\relax]{figures/assets/#1}};
        \begin{scope}[x={(img.south east)},y={(img.north west)}]
          \clip (0,0) rectangle (1,1);
          #2
        \end{scope}
      \end{tikzpicture}}}
  \newcommand{\pvimgmirrorann}[2]{%
    \parbox[c][\dimexpr\pvcellheight+2\fboxsep+2\fboxrule\relax][c]{%
        \dimexpr\pvcellwidth+2\fboxsep+2\fboxrule\relax}{%
      \begin{tikzpicture}
        \node[anchor=south west,inner sep=0] (img) at (0,0) {%
          \reflectbox{\includegraphics[
            width=\dimexpr\pvcellwidth+2\fboxsep+2\fboxrule\relax,
            height=\dimexpr0.75\pvcellwidth+1.5\fboxsep+1.5\fboxrule\relax]{figures/assets/#1}}};
        \begin{scope}[x={(img.south east)},y={(img.north west)}]
          \clip (0,0) rectangle (1,1);
          #2
        \end{scope}
      \end{tikzpicture}}}
  \tikzset{
    pvbox/.style={draw=red,line width=0.45pt},
    pvarrow/.style={draw=red,line width=0.9pt,
      -{Stealth[length=1.6mm,width=1.3mm]}},
    pvtag/.style={anchor=south west,fill=red,fill opacity=0.7,text opacity=1,
      text=white,rounded corners=0.3mm,inner xsep=0.8pt,inner ysep=0.5pt,
      font=\fontsize{3.4}{3.8}\selectfont\bfseries}}
  \newcommand{\pvmark}[1]{{\bfseries\color{red}#1}}

  \begin{tabular}{@{}c@{\hspace{1mm}}c@{\hspace{1mm}}c@{\hspace{3mm}}c@{\hspace{1mm}}c@{\hspace{1mm}}c@{}}
    \multicolumn{3}{c}{\scriptsize Early turn}
    & \multicolumn{3}{c}{\scriptsize Later turn} \\[0.3mm]
    {\scriptsize Left wrist} & {\scriptsize Head} & {\scriptsize Right wrist}
    & {\scriptsize Left wrist} & {\scriptsize Head} & {\scriptsize Right wrist} \\[0.4mm]
    \pvimg{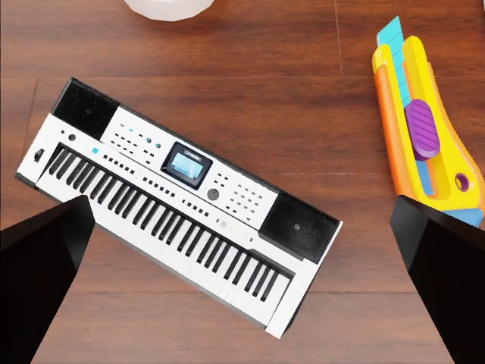}
    & \pvimgmirrorann{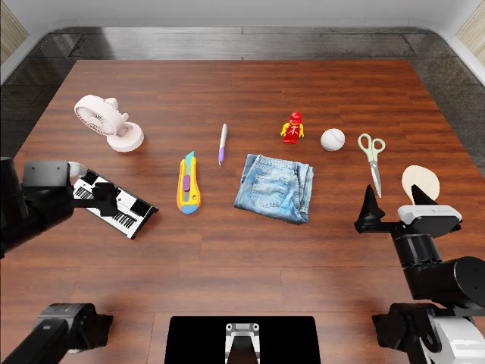}{%
        \draw[pvbox] (0.76,0.40) rectangle (0.99,0.58);
        \node[pvtag,anchor=north east] at (0.99,0.385) {wrong arm};}
    & \pvimg{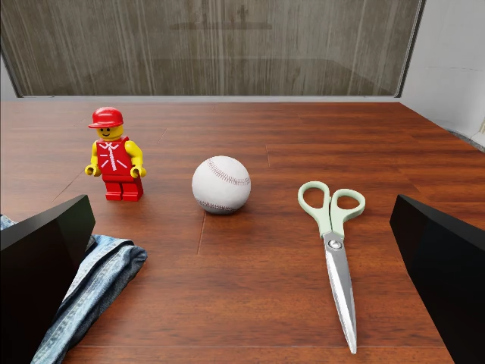}
    & \pvimg{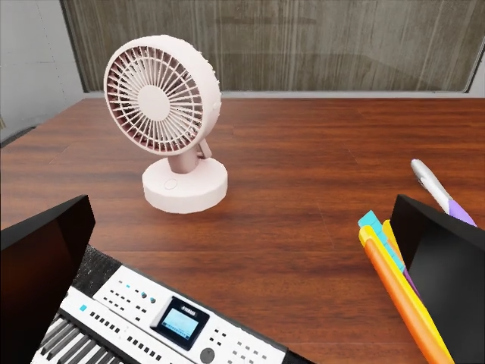}
    & \pvimgmirrorann{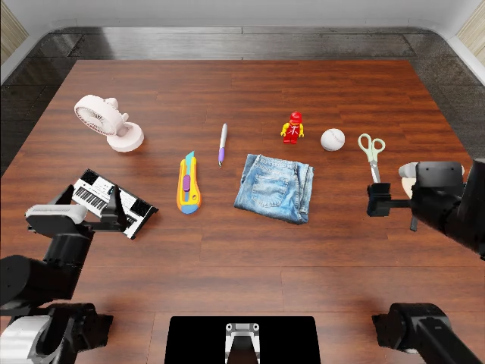}{%
        \draw[pvbox] (0.015,0.375) rectangle (0.225,0.575);
        \node[pvtag,anchor=north west] at (0.015,0.36) {correct arm};}
    & \pvimg{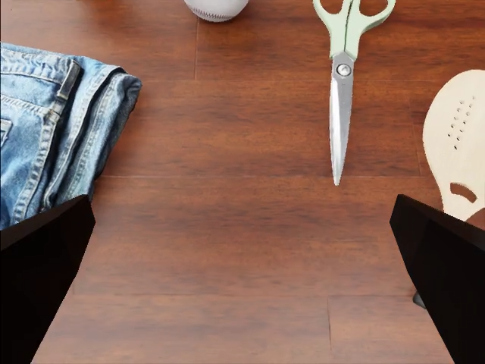} \\[0.4mm]
    \multicolumn{6}{c}{\scriptsize (a) Left--right image mirror \;\textemdash\;
      \texttt{general\_pickup} layout 0, pick up the scissors} \\[0.6mm]
    \multicolumn{6}{c}{\parbox{0.96\linewidth}{\centering\scriptsize
      Reasoning written by the model between the two turns:
      \itshape ``The camera views show that the \pvmark{right arm} is the one
      beside the scissors. I am restoring the idle \pvmark{left arm} and
      preparing the \pvmark{right hand} for a downward grasp.''}} \\[1.4mm]
    \pvcell{masked} & \pvcell{masked}
    & \pvimgann{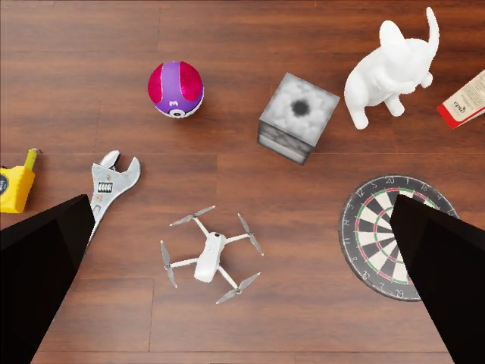}{%
        \node[pvtag] at (0.02,0.62) {searching};}
    & \pvcell{masked} & \pvcell{masked}
    & \pvimgann{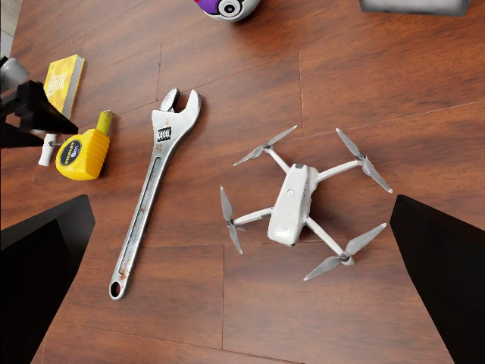}{%
        \draw[pvbox] (0.055,0.515) rectangle (0.135,0.66);
        \node[pvtag,anchor=north west,align=left] at (0.01,0.50)
          {found\\left-hand grasp};} \\[0.4mm]
    \multicolumn{6}{c}{\scriptsize (b) Right-wrist view only \;\textemdash\;
      \texttt{general\_pickup} layout 9} \\[0.6mm]
    \multicolumn{6}{c}{\parbox{0.96\linewidth}{\centering\scriptsize
      The model sweeps the right arm to search for the object with its wrist
      camera, then grasps with the left hand.}} \\[1.4mm]
    \pvimg{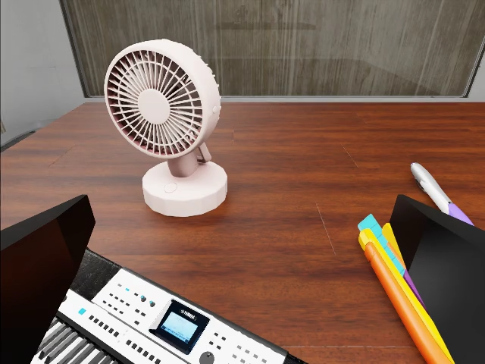}
    & \pvimgann{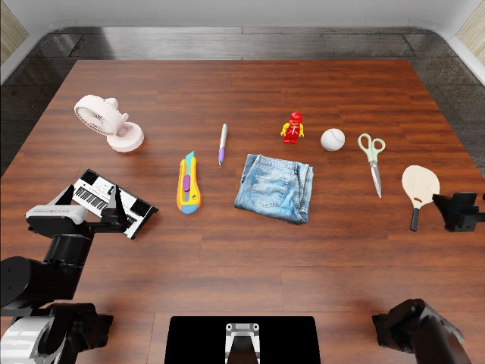}{%
        \draw[pvarrow] (0.80,0.50) -- (0.98,0.50);
        \node[pvtag,anchor=south east] at (0.82,0.28) {perturbed};}
    & \pvimg{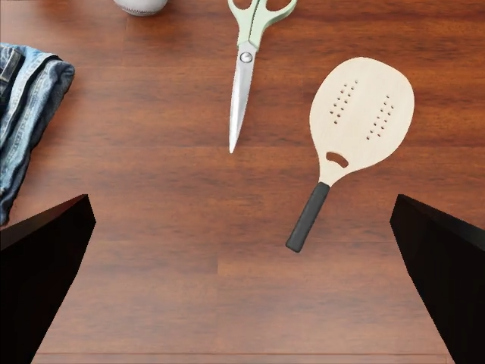}
    & \pvimg{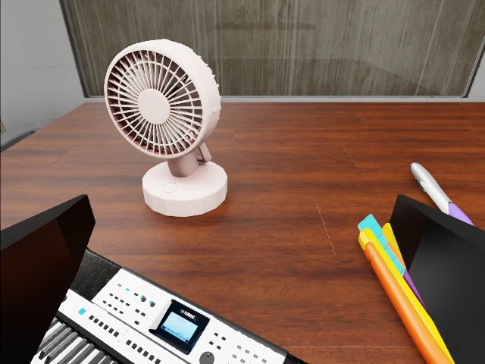}
    & \pvimgann{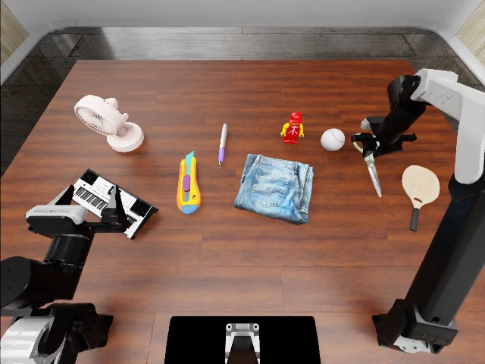}{%
        \draw[pvbox] (0.705,0.435) rectangle (0.805,0.675);
        \node[pvtag,anchor=north east] at (0.80,0.37) {Target reach};}
    & \pvimg{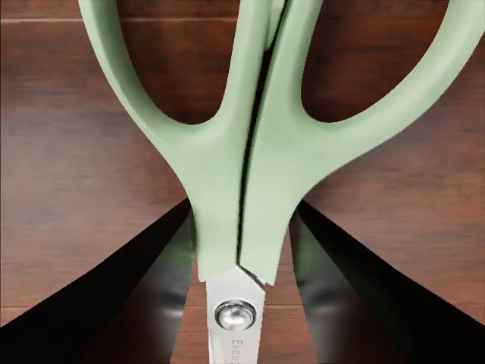} \\[0.4mm]
    \multicolumn{6}{c}{\scriptsize (c) Per-move pose jitter \;\textemdash\;
      \texttt{general\_pickup} layout 0} \\[0.6mm]
    \multicolumn{6}{c}{\parbox{0.96\linewidth}{\centering\scriptsize
      Every executed move is perturbed, and the model corrects until the grasp
      succeeds.}} \\
  \end{tabular}

  \caption{The observations the model receives under three perturbations, defined in
  Appendix~\ref{sec:app-perturbation-probes}. Each row is one condition, named under its
  frames; the left group of three views is an early turn and the right group a later turn
  of the same episode. The line under row (a) is the model's own reasoning from the
  episode, and the lines under rows (b) and (c) describe an observed behaviour instead.
  Row (a) shows real frames from the \texttt{flip\_vision\_lr} episode on layout 0. The
  mirror reverses apparent left--right positions without changing camera identities.
  Under (b) only the right wrist camera is available, and the masked
  cells mark the views the model does not receive. Row (b) comes from the supplementary
  rerun with 170 calls and 400 environment steps; the standard-budget run failed.
  Under (c) every executed move is
  offset by a random displacement of mean magnitude 10\,cm. The model receives
  both images and motion feedback; their contributions are not isolated
  (Appendix~\ref{sec:app-perturbation-traces}).}
  \label{fig:icl-perturbation}
\end{figure}
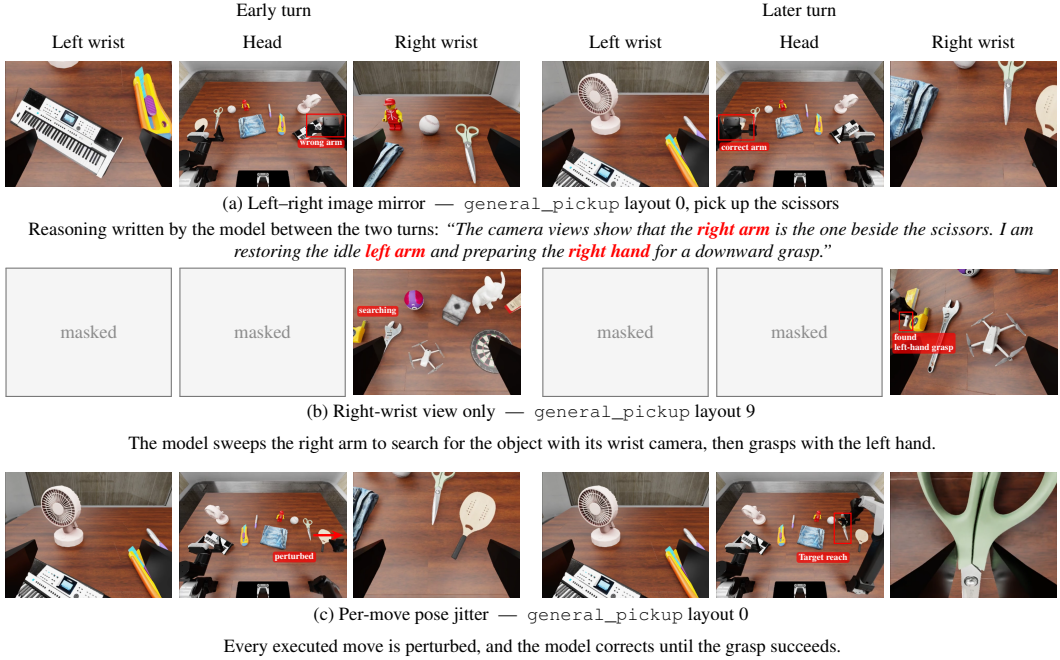

Figure~\ref{fig:icl-perturbation} shows what the model receives under three of these
conditions, at an early and a later turn of the same episode. In row (a) the model
writes down, between the two turns, why it moves the grasp to the other hand. That note
documents a change in the model's stated arm assignment.
Row (c) shows the same loop under pose jitter: an early command lands far to the right
of the object, and later attempts lead to a successful grasp. Row (b) is a
supplementary larger-budget rerun, not a standard-budget success in Table~\ref{tab:f4}.

Under Negated Cartesian axes, one successful episode includes a note that the
model's forward/back estimate was reversed. The note explicitly refers to the
wrist view. Table~\ref{tab:a6} reproduces selected calls from that episode.
These observations are consistent with within-episode adaptation, but do not
identify which inputs caused the correction or establish complete identification
of the coordinate mapping.

\section{LLM as policy on real hardware}
\label{sec:real}

We did not complete the official RoboDojo-Real protocol with GPT-6
Astra~\citep{robodojo2026}. Testing was halted after the model repeatedly issued
physically unreasonable or unsafe actions, including incidents that damaged
hardware; no person was injured. The hardware material is therefore a selective
sample rather than an official result, and it cannot establish Astra's general
real-world reliability.

We keep the remaining trials as diagnostic material. They suggest that Astra
interpreted the tasks more reliably than it executed them safely, which is a
qualitative observation rather than a benchmark result. Appendix~\ref{sec:app-real}
reports the retained trials, their scores, the embodiments, and the separate
joint-position deployments. Section~\ref{sec:beyond} reports two further evaluations
of Astra that also fall outside the official RoboDojo protocol.

\section{Beyond tabletop manipulation}
\label{sec:beyond}

The official RoboDojo suite is tabletop manipulation. We also ran two evaluations of
GPT-6 Astra outside that setting. Neither is part of the official RoboDojo protocol,
and neither produces a Score that belongs on the Section~\ref{sec:results-sim} board.
The humanoid recordings are qualitative; the piano runs include task-specific
scores. Neither establishes a general performance level beyond the tested runs.
\\[2pt]
\textbf{Mobile humanoid grasping.} Astra was also deployed through a separate
mobile-humanoid interface. The recordings show walking followed by grasping, and
a longer sequence of walking, sitting, and picking while seated. No success rate
or process score was computed for either. Appendix~\ref{sec:app-demos} records the
action vocabulary, horizons, and clips. We do not use these recordings to
establish compliance with the strict LLM-as-policy definition in Section~\ref{sec:study-boundary}.
\\[2pt]
\textbf{Bimanual piano playing.} A second evaluation asks Astra to play on an
88-key piano. Here it does not emit one motion target per turn. It writes a controller
that outputs 45-D joint commands every $0.05$\,s for two Shadow hands. The LLM's
weights remain fixed, but it refines its code through simulation trials without
access to a reference solution. These runs use a piano-performance F1 metric,
which is not comparable with a RoboDojo Score. Appendix~\ref{sec:app-piano}
reports the isolation rules, practice runs, and whether the delivered program survives
a change of music.

\section{Conclusion}
\label{sec:conclusion}

This study evaluates LLM as policy: a language model selects robot motion targets
without a pretrained motor policy in the control path. Across 42 RoboDojo tasks,
GPT-6 Astra achieves 28.97 Score and 22.48\% average SR, ranking above the 40
public policies in our comparison. GPT-5.5 and DeepSeek-Flash perform substantially
worse with the same interface, so the result does not generalize to LLMs as a
class. Astra's lead on Open and Generalization coexists with large gaps in
precision and coordinated manipulation. Unsafe actions halted our real-robot
trials. Aggregate benchmark performance therefore does not establish reliable or
safe physical control.

The context experiments suggest that learning from demonstrations and correcting
actions during execution should be evaluated separately. A single demonstration
from another layout reduces aggregate success in our matched trials. Yet under
visual and action perturbations, selected episodes show Astra revising spatial
judgments and later actions after seeing the outcome of a motion. These traces
are preliminary evidence of within-episode adaptation. A next step is to test
whether such corrections transfer across tasks and embodiments. Turning them
into reliable, precise execution remains a central challenge for LLMs used as
policies.

\label{marker:endofmaintext}

\section*{Acknowledgements}

We thank Mengdi Xu for suggestions and for providing the setup for the
real-robot experiments.

\bibliographystyle{iclr2026_conference}
\bibliography{refs}

\begin{thebibliography}{44}
\providecommand{\natexlab}[1]{#1}
\providecommand{\url}[1]{\texttt{#1}}
\expandafter\ifx\csname urlstyle\endcsname\relax
  \providecommand{\doi}[1]{doi: #1}\else
  \providecommand{\doi}{doi: \begingroup \urlstyle{rm}\Url}\fi

\bibitem[Ahn et~al.(2023)]{saycan}
Michael Ahn et~al.
\newblock {Do As I Can, Not As I Say: Grounding Language in Robotic
  Affordances}.
\newblock In \emph{{Proc. CoRL}}, volume 205 of \emph{PMLR}, 2023.
\newblock URL \url{https://proceedings.mlr.press/v205/ichter23a.html}.

\bibitem[Bajcsy(1988)]{bajcsy1988active}
Ruzena Bajcsy.
\newblock {Active Perception}.
\newblock \emph{Proceedings of the IEEE}, 76\penalty0 (8):\penalty0 966--1005,
  1988.

\bibitem[Battaglia et~al.(2013)Battaglia, Hamrick, and
  Tenenbaum]{battaglia2013simulation}
Peter~W. Battaglia, Jessica~B. Hamrick, and Joshua~B. Tenenbaum.
\newblock {Simulation as an Engine of Physical Scene Understanding}.
\newblock \emph{Proceedings of the National Academy of Sciences}, 110\penalty0
  (45):\penalty0 18327--18332, 2013.

\bibitem[Berman et~al.(2026)Berman, Ilie, Deng, and
  Freeman]{berman2026robotics}
Shmuel Berman, Michael Ilie, Jia Deng, and Daniel Freeman.
\newblock {Claude Plays Robotics}.
\newblock Anthropic Research, July 2026.
\newblock URL \url{https://www.anthropic.com/research/claude-plays-robotics}.

\bibitem[Black et~al.(2024)]{pi0}
Kevin Black et~al.
\newblock {$\pi_0$: A Vision-Language-Action Flow Model for General Robot
  Control}.
\newblock \emph{arXiv preprint arXiv:2410.24164}, 2024.
\newblock URL \url{https://arxiv.org/abs/2410.24164}.

\bibitem[Brohan et~al.(2022)]{rt1}
Anthony Brohan et~al.
\newblock {RT-1: Robotics Transformer for Real-World Control at Scale}.
\newblock \emph{arXiv preprint arXiv:2212.06817}, 2022.
\newblock URL \url{https://arxiv.org/abs/2212.06817}.

\bibitem[Brohan et~al.(2023)]{rt2}
Anthony Brohan et~al.
\newblock {RT-2: Vision-Language-Action Models Transfer Web Knowledge to
  Robotic Control}.
\newblock In \emph{{Proc. CoRL}}, volume 229 of \emph{PMLR}, 2023.
\newblock URL \url{https://arxiv.org/abs/2307.15818}.

\bibitem[Brown et~al.(2020)]{brown2020language}
Tom Brown et~al.
\newblock {Language Models are Few-Shot Learners}.
\newblock In \emph{{Adv.\ Neural Inf.\ Process.\ Syst.}}, 2020.
\newblock URL \url{https://arxiv.org/abs/2005.14165}.

\bibitem[Chen et~al.(2026)Chen, Wang, Cui, Zhou, Shao, Li, Su, Gan, Wang, Fu,
  Yang, and Yue]{host}
Guangyan Chen, Meiling Wang, Te~Cui, Zichen Zhou, Qi~Shao, Xiaofan Li, Hang Su,
  Ruyi Gan, Hao Wang, Mengyin Fu, Yi~Yang, and Yufeng Yue.
\newblock {Robots Acquire Manipulation Skills in Seconds from a Single Human
  Video}.
\newblock \emph{arXiv preprint arXiv:2607.20033}, 2026.
\newblock URL \url{https://arxiv.org/abs/2607.20033}.

\bibitem[Chi et~al.(2023)]{diffusionpolicy}
Cheng Chi et~al.
\newblock {Diffusion Policy: Visuomotor Policy Learning via Action Diffusion}.
\newblock In \emph{{Proc. RSS}}, 2023.
\newblock URL \url{https://arxiv.org/abs/2303.04137}.

\bibitem[Community et~al.(2026)Community, Chen, Chen, Nian,
  et~al.]{community2026xpolicylab}
XPolicyLab Community, Tianxing Chen, Yue Chen, Tian Nian, et~al.
\newblock {XPolicyLab: A Unified Standard and Open Ecosystem for Robot Policy
  Evaluation and Deployment}.
\newblock \emph{arXiv preprint arXiv:2608.09892}, 2026.
\newblock URL \url{https://arxiv.org/abs/2608.09892}.

\bibitem[Driess et~al.(2023)]{palme}
Danny Driess et~al.
\newblock {PaLM-E: An Embodied Multimodal Language Model}.
\newblock In \emph{{Proc. ICML}}, volume 202 of \emph{PMLR}, 2023.
\newblock URL \url{https://arxiv.org/abs/2303.03378}.

\bibitem[{Figure AI}(2025)]{helix}
{Figure AI}.
\newblock {Helix: A Vision-Language-Action Model for Generalist Humanoid
  Control}.
\newblock Technical report, February 2025.
\newblock URL \url{https://www.figure.ai/news/helix}.

\bibitem[Fu et~al.(2024)Fu, Huang, Datta, Chen, Panitch, Liu, Li, and
  Goldberg]{fu2024icrt}
Max~Letian Fu, Huang Huang, Gaurav Datta, Lawrence~Yunliang Chen, Will Panitch,
  Fangchen Liu, Hui Li, and Ken Goldberg.
\newblock {In-Context Imitation Learning via Next-Token Prediction}.
\newblock \emph{arXiv preprint arXiv:2408.15980}, 2024.
\newblock URL \url{https://arxiv.org/abs/2408.15980}.

\bibitem[{Generalist}(2026)]{gen15}
{Generalist}.
\newblock {GEN-1.5: Embodied Foundation Models are One-Shot Learners}.
\newblock Technical report, August 2026.
\newblock URL \url{https://generalistai.com/blog/gen-1.5}.

\bibitem[Ghosh et~al.(2024)]{octo}
Dibya Ghosh et~al.
\newblock {Octo: An Open-Source Generalist Robot Policy}.
\newblock In \emph{{Proc. RSS}}, 2024.
\newblock URL \url{https://roboticsproceedings.org/rss20/p090.html}.

\bibitem[Huang et~al.(2023{\natexlab{a}})Huang, Wang, Zhang, Li, Wu, and
  Fei-Fei]{voxposer}
Wenlong Huang, Chen Wang, Ruohan Zhang, Yunzhu Li, Jiajun Wu, and Li~Fei-Fei.
\newblock {VoxPoser: Composable 3D Value Maps for Robotic Manipulation with
  Language Models}.
\newblock In \emph{{Proc. CoRL}}, volume 229 of \emph{PMLR}, pp.\  540--562,
  2023{\natexlab{a}}.
\newblock URL \url{https://proceedings.mlr.press/v229/huang23b.html}.

\bibitem[Huang et~al.(2023{\natexlab{b}})]{innermonologue}
Wenlong Huang et~al.
\newblock {Inner Monologue: Embodied Reasoning through Planning with Language
  Models}.
\newblock In \emph{{Proc. CoRL}}, volume 205 of \emph{PMLR},
  2023{\natexlab{b}}.
\newblock URL \url{https://proceedings.mlr.press/v205/huang23c.html}.

\bibitem[Ilie et~al.(2026)Ilie, Freeman, and Troy]{projectfetch2026}
Michael Ilie, C.~Daniel Freeman, and Kevin~K. Troy.
\newblock {Project Fetch: Phase Two}.
\newblock Anthropic Research, June 2026.
\newblock URL \url{https://www.anthropic.com/research/project-fetch-phase-two}.

\bibitem[Jia et~al.(2026)Jia, Lin, Zhang, Zhang, Liu, and
  Jiang]{agentaspolicy2026}
Mengzhao Jia, Yang Lin, Xixin Zhang, Zhihan Zhang, Xiaobai Liu, and Meng Jiang.
\newblock {Agent as Policy for Robotic Manipulation}.
\newblock \emph{arXiv preprint arXiv:2609.12541}, 2026.
\newblock URL \url{https://arxiv.org/abs/2609.12541}.

\bibitem[Jiang et~al.(2026)Jiang, Chebotar, Zheng, Hu, Ge, Wu, Dai, Reed,
  Fei-Fei, Zhu, and Fan]{robottt}
Yunfan Jiang, Yevgen Chebotar, Ruijie Zheng, Fengyuan Hu, Yunhao Ge, Jimmy Wu,
  Tianyuan Dai, Scott Reed, Li~Fei-Fei, Yuke Zhu, and Linxi~Jim Fan.
\newblock {RoboTTT: Context Scaling for Robot Policies}.
\newblock \emph{arXiv preprint arXiv:2607.15275}, 2026.
\newblock URL \url{https://arxiv.org/abs/2607.15275}.

\bibitem[Kahneman(2011)]{kahneman2011thinking}
Daniel Kahneman.
\newblock \emph{{Thinking, Fast and Slow}}.
\newblock Farrar, Straus and Giroux, 2011.

\bibitem[Kim et~al.(2026)Kim, Gao, Lin, Lin, Ge, Lam, Liang, Song, Liu, Finn,
  and Gu]{cosmospolicy}
Moo~Jin Kim, Yihuai Gao, Tsung-Yi Lin, Yen-Chen Lin, Yunhao Ge, Grace Lam,
  Percy Liang, Shuran Song, Ming-Yu Liu, Chelsea Finn, and Jinwei Gu.
\newblock {Cosmos Policy: Fine-Tuning Video Models for Visuomotor Control and
  Planning}.
\newblock \emph{arXiv preprint arXiv:2601.16163}, 2026.
\newblock URL \url{https://arxiv.org/abs/2601.16163}.

\bibitem[Kim et~al.(2025)]{openvla}
Moo~Jin Kim et~al.
\newblock {OpenVLA: An Open-Source Vision-Language-Action Model}.
\newblock In \emph{{Proc. CoRL}}, volume 270 of \emph{PMLR}, 2025.
\newblock URL \url{https://proceedings.mlr.press/v270/kim25c.html}.

\bibitem[Kwon et~al.(2024)Kwon, Di~Palo, and Johns]{kwon2024trajectory}
Teyun Kwon, Norman Di~Palo, and Edward Johns.
\newblock {Language Models as Zero-Shot Trajectory Generators}.
\newblock \emph{IEEE Robotics and Automation Letters}, 2024.
\newblock \doi{10.1109/LRA.2024.3410155}.
\newblock URL \url{https://arxiv.org/abs/2310.11604}.

\bibitem[Liang et~al.(2023)]{codeaspolicies}
Jacky Liang et~al.
\newblock {Code as Policies: Language Model Programs for Embodied Control}.
\newblock In \emph{{Proc. ICRA}}, 2023.
\newblock URL \url{https://arxiv.org/abs/2209.07753}.

\bibitem[Liu et~al.(2023)Liu, Zhu, Gao, Feng, Liu, Zhu, and
  Stone]{liu2023libero}
Bo~Liu, Yifeng Zhu, Chongkai Gao, Yihao Feng, Qiang Liu, Yuke Zhu, and Peter
  Stone.
\newblock {LIBERO: Benchmarking Knowledge Transfer for Lifelong Robot
  Learning}.
\newblock In \emph{{Adv.\ Neural Inf.\ Process.\ Syst.}}, 2023.
\newblock URL \url{https://arxiv.org/abs/2306.03310}.

\bibitem[Liu et~al.(2025)Liu, Pathak, and Agarwal]{locoformer}
Min Liu, Deepak Pathak, and Ananye Agarwal.
\newblock {LocoFormer: Generalist Locomotion via Long-context Adaptation}.
\newblock In \emph{{Proc. CoRL}}, volume 305 of \emph{PMLR}, pp.\  532--546,
  2025.
\newblock URL \url{https://proceedings.mlr.press/v305/liu25a.html}.

\bibitem[Min et~al.(2022)]{min2022rethinking}
Sewon Min et~al.
\newblock {Rethinking the Role of Demonstrations: What Makes In-Context
  Learning Work?}
\newblock In \emph{{Proc.\ EMNLP}}, 2022.
\newblock URL \url{https://arxiv.org/abs/2202.12837}.

\bibitem[{Open X-Embodiment Collaboration} et~al.(2024)]{openx}
{Open X-Embodiment Collaboration} et~al.
\newblock {Open X-Embodiment: Robotic Learning Datasets and RT-X Models}.
\newblock In \emph{{Proc. ICRA}}, 2024.
\newblock URL \url{https://arxiv.org/abs/2310.08864}.

\bibitem[{OpenAI}(2023)]{gpt4}
{OpenAI}.
\newblock {GPT-4 Technical Report}.
\newblock \emph{arXiv preprint arXiv:2303.08774}, 2023.
\newblock URL \url{https://arxiv.org/abs/2303.08774}.

\bibitem[{OpenAI}(2026)]{gpt6astra}
{OpenAI}.
\newblock {GPT-6 Astra: A New Generation of Intelligence}.
\newblock OpenAI, September 2026.
\newblock URL \url{https://openai.com/index/gpt-6-astra/}.

\bibitem[{Robocurve}(2026)]{robocurveastra2026}
{Robocurve}.
\newblock {GPT-6 Astra on Robotic Manipulation}.
\newblock Robocurve report, September 2026.
\newblock URL \url{https://openai.robocurve.org/gpt-6-astra/}.

\bibitem[{RoboDojo Team}(2026)]{robodojo2026}
{RoboDojo Team}.
\newblock {RoboDojo: A Unified Sim-and-Real Benchmark for Comprehensive
  Evaluation of Generalist Robot Manipulation Policies}.
\newblock \emph{arXiv preprint arXiv:2607.04434}, 2026.
\newblock URL \url{https://arxiv.org/abs/2607.04434}.

\bibitem[Shi et~al.(2025)]{hirobot}
Lucy~Xiaoyang Shi et~al.
\newblock {Hi Robot: Open-Ended Instruction Following with Hierarchical
  Vision-Language-Action Models}.
\newblock In \emph{{Proc. ICML}}, volume 267 of \emph{PMLR}, pp.\
  54919--54933, 2025.
\newblock URL \url{https://proceedings.mlr.press/v267/shi25d.html}.

\bibitem[{Skild AI}(2026)]{skilds1}
{Skild AI}.
\newblock {Introducing S1: In-Context Learning for Robotics}.
\newblock Technical report, August 2026.
\newblock URL \url{https://skild.ai/blogs/s1}.

\bibitem[Tsui et~al.(2026)Tsui, Fang, and Hwu]{faea2026}
Brian~Y. Tsui, Alan~Y. Fang, and Tiffany~J. Hwu.
\newblock {Demonstration-Free Robotic Control via LLM Agents}.
\newblock \emph{arXiv preprint arXiv:2601.20334}, 2026.
\newblock URL \url{https://arxiv.org/abs/2601.20334}.

\bibitem[Wang et~al.(2026)Wang, Shi, Fei, Fu, Ji, Gong, and Qiu]{icwm}
Siyin Wang, Junhao Shi, Senyu Fei, Zhaoyang Fu, Li~Ji, Jingjing Gong, and
  Xipeng Qiu.
\newblock {In-Context World Modeling for Robotic Control}.
\newblock \emph{arXiv preprint arXiv:2606.26025}, 2026.
\newblock URL \url{https://arxiv.org/abs/2606.26025}.

\bibitem[Ye et~al.(2026)]{dreamzero}
Seonghyeon Ye et~al.
\newblock {World Action Models are Zero-shot Policies}.
\newblock \emph{arXiv preprint arXiv:2602.15922}, 2026.
\newblock URL \url{https://arxiv.org/abs/2602.15922}.

\bibitem[Yin et~al.(2024)Yin, Wang, Sharma, Niu, Darrell, and
  Herzig]{yin2024roboprompt}
Yida Yin, Zekai Wang, Yuvan Sharma, Dantong Niu, Trevor Darrell, and Roei
  Herzig.
\newblock {In-Context Learning Enables Robot Action Prediction in LLMs}.
\newblock \emph{arXiv preprint arXiv:2410.12782}, 2024.
\newblock URL \url{https://arxiv.org/abs/2410.12782}.

\bibitem[Zakka et~al.(2023)]{robopianist}
Kevin Zakka et~al.
\newblock {RoboPianist: Dexterous Piano Playing with Deep Reinforcement
  Learning}.
\newblock In \emph{{Proc. CoRL}}, volume 229 of \emph{PMLR}, pp.\  2975--2994,
  2023.
\newblock URL \url{https://proceedings.mlr.press/v229/zakka23a.html}.

\bibitem[Zeng \& {the Generalist Team}(2026)Zeng and {the Generalist
  Team}]{physicalcommonsense}
Andy Zeng and {the Generalist Team}.
\newblock {The Dark Matter of Robotics: Physical Commonsense}.
\newblock Generalist blog, January 2026.
\newblock URL \url{https://generalistai.com/blog/physical-commonsense}.

\bibitem[Zhang et~al.(2026)Zhang, Li, Yang, Chen, Liu, Liu, and Ma]{tttvla}
Wenbo Zhang, Jianxiong Li, Shuai Yang, Sijin Chen, Jiajun Liu, Lingqiao Liu,
  and Xiao Ma.
\newblock {TTT-VLA: Test-Time Latent Prompt Optimization for
  Vision-Language-Action Models}.
\newblock \emph{arXiv preprint arXiv:2606.03127}, 2026.
\newblock URL \url{https://arxiv.org/abs/2606.03127}.

\bibitem[Zhao et~al.(2023)]{act}
Tony~Z. Zhao et~al.
\newblock {Learning Fine-Grained Bimanual Manipulation with Low-Cost Hardware}.
\newblock In \emph{{Proc. RSS}}, 2023.
\newblock URL \url{https://arxiv.org/abs/2304.13705}.

\end{thebibliography}

\appendix
\section{Prompt and tool interface}
\label{sec:app-prompt}

This appendix makes the model-facing interface explicit. An episode begins with
one system message and one Goal message. Every control turn then appends one
multimodal observation, one assistant tool call, and one tool result. All text
remains in context. Images remain only on the two most recent observation turns.

\paragraph{System message.} The header below is abridged only where marked. The
omitted embodiment notes contain task-independent geometry and operating facts:
world axes, arm bases, table height, reach, gripper polarity, active-view advice,
contact-step advice, and the requirement to return both arms to their starting
poses when the benchmark scores it.

\begin{small}
\begin{verbatim}
You are controlling a real robot embodiment named 'robodojo-arx-x5'.
You receive RGB camera images, the current world-frame grasp-point
state of both arms in the same 14 dimensions move_eef takes, arm joint
angles as context you cannot command, and a task instruction.
Move with move_eef by naming only the world-frame dimensions you want
to change. Cartesian targets must be estimated from RGB; no depth or
world-coordinate query is available. Respond with exactly one tool
call per turn. After each motion the next observation reports how far
the grasp point ended up from what you asked for, so check it before
assuming a motion landed. Two budgets run down at once and whichever
empties first ends the episode: 100 LLM calls, one per turn, and the
environment's own step limit, reported with each observation as the
env steps remaining. A motion spends env steps in proportion to how
far it travels, so a small correction is cheap and a long reach is not.

Embodiment notes:
<task-independent robot geometry and operating notes>
\end{verbatim}
\end{small}

\paragraph{Goal message.} The following shows the task-level content for
\texttt{general\_pickup}, apart from the episode-specific instruction shown as a
placeholder. All 42 tasks use the same template and a task-specific wiki recipe.

\begin{small}
\begin{verbatim}
Goal: <official episode instruction>

TASK RECIPE:
# General Pickup

Official RoboDojo wiki capability dimension, Description, and
process-score ladder. The live Goal instruction is still authoritative
for instance-specific slots. RoboDojo's reward judges the episode;
these rows are the environment's partial-credit scores, not a
substitute for official success.

## Capability dimension
Open -- Open-ended or language/image-conditioned manipulation tasks.

## Description
There are multiple objects. The robot needs to understand the language
instruction, identify the target object, and pick it up.

## Scoring
0: The target object is not lifted high enough.
100: The target object is lifted at least 10 cm.
\end{verbatim}
\end{small}

\paragraph{Interaction history.} At turn $t$, the API receives the system and Goal
messages followed by the alternating observations, assistant tool calls, and tool
results from turns $1,\ldots,t-1$. The current observation is last. When
$\texttt{image\_horizon}=2$, images in older observation messages are replaced by
the text \texttt{[earlier camera image omitted to save context]}; their state text
is unchanged.

\paragraph{Observation message.} Values and camera pixels change each turn; the
field structure does not. The 14 grasp-point values are absolute targets in the
same coordinate system used by \texttt{move\_eef}. Joint angles are included only
as read-only proprioception.

\begin{small}
\begin{verbatim}
Instruction: <official episode instruction>
World-frame grasp-point state (metres; degrees from straight down):
left_x=<v>  left_y=<v>  left_z=<v>
left_pitch_deg=<v>  left_roll_deg=<v>  left_yaw_deg=<v>
left_gripper=<v>
right_x=<v>  right_y=<v>  right_z=<v>
right_pitch_deg=<v>  right_roll_deg=<v>  right_yaw_deg=<v>
right_gripper=<v>
Arm joint angles, radians, for context; not commandable:
left_joint1=<v> ... left_joint6=<v>
right_joint1=<v> ... right_joint6=<v>
Arrival check for the previous move_eef target: <optional error>.
Env steps remaining before the episode ends: <n>
camera 'head' (step <t>): <RGB JPEG>
camera '<left wrist>' (step <t>): <RGB JPEG>
camera '<right wrist>' (step <t>): <RGB JPEG>
\end{verbatim}
\end{small}

\paragraph{Tools.} The model must return exactly one of the following calls.
The \texttt{targets} object must be nonempty, but may contain any subset of the
14 dimensions. Unnamed dimensions hold their observed values.

\begin{small}
\begin{verbatim}
move_eef({
  "targets": {"<dimension>": <number>, ...},
  "note": "<current observation and reason for this motion>"
})

give_up({
  "reason": "<why the episode is unrecoverable>",
  "hindsight": "<what would have been needed>"
})
\end{verbatim}
\end{small}

Position units are metres. The position bounds are
\texttt{left\_x}$\in[-1.1,0.5]$,
\texttt{right\_x}$\in[-0.5,1.1]$,
\texttt{left/right\_y}$\in[-1.25,0.35]$, and
\texttt{left/right\_z}$\in[0.7,1.565]$.
For either arm,
\texttt{pitch\_deg}$\in[-180,180]$,
\texttt{roll\_deg}$\in[-90,90]$,
\texttt{yaw\_deg}$\in[-180,180]$, and
\texttt{gripper}$\in[0,1]$, where 0 is closed and 1 is open. Zero orientation
denotes a straight-down grasp. An invalid or unreachable request returns an
error as the tool result and does not move the robot.

\paragraph{Tool result and next observation.} A valid call first adds a result of
the following form to the textual history:

\begin{small}
\begin{verbatim}
Accepted: playing <N> waypoints (<seconds>s).
The next observation reports how close the grasp point landed.
\end{verbatim}
\end{small}

An invalid target instead returns a specific error, such as
\texttt{target for 'left\_z' is outside [0.7, 1.565]}, and leaves the robot
stationary while the model chooses another target. After an accepted chunk is
executed, RoboDojo supplies the next three RGB images, robot state, and remaining
step budget. The interface formats these as the next observation message and adds
an arrival check such as \texttt{left 20 mm and 0.0 deg away}. Thus the tool result
reports whether the request was executable, while the next observation reports what
physically happened.

\section{Full simulation boards}
\label{sec:app-boards}

Table~\ref{tab:sim-board} is the complete 43-row RoboDojo-Sim board that
Table~\ref{tab:board-compact} excerpts. Table~\ref{tab:per-task} is the official
Per-Task board plus the two LLM-controller columns from the 50-episode
single-seed dump and the DeepSeek-Flash \dag{} column from the 10-episode
campaign; Generalization cells are the mean of Gen-Std and Gen-Rand. The
complementary task subsets are reported in Table~\ref{tab:f2} in the main text.

DeepSeek-Flash records successes on four tasks: \texttt{general\_pickup}
(30\% SR), \texttt{stack\_blocks} (20\%), \texttt{align\_blocks} (20\%), and
\texttt{match\_and\_pick\_from\_conveyor} (10\%). These SR values differ from
the process-reward Scores in Table~\ref{tab:per-task}.

\begin{table}[p]
  \centering
  \caption{RoboDojo-Sim board in the official Score/SR\% cell format: the 40 public
  policies plus GPT-6 Astra, DeepSeek-Flash, and GPT-5.5, sorted by Average Score and
  ranked 1--43 together. Ranking the three LLM controllers alongside the board makes
  the comparison legible; the public submission itself contains only the 40 policy rows,
  where the same order gives DM0.5 rank~1. \dag~DeepSeek-Flash is 1 seed
  $\times$ 10 episodes per task, not 50. Snapshot of the policy rows taken 2026-09-10
  from the public leaderboard.}
  \label{tab:sim-board}
  \small
  \setlength{\tabcolsep}{4pt}
  \resizebox{\linewidth}{!}{%
  \begin{tabular}{rlcccccc}
    \toprule
    Rank & Model & Average & Gen. & Prec. & Long-Hor. & Memory & Open \\
    \midrule
    1 & \textbf{GPT-6 Astra} & 28.97/22.48 & 33.36/30.50 & 12.65/4.00 & 21.45/8.25 & 43.04/38.67 & 34.36/31.00 \\
    2 & DM0.5 & 24.90/19.34 & 15.77/10.95 & 24.82/16.75 & 33.70/19.50 & 47.74/47.44 & 2.43/2.08 \\
    3 & GalaxeaVLA (G0.5) & 20.23/14.88 & 18.46/12.83 & 28.25/20.42 & 44.12/32.25 & 8.61/7.33 & 1.73/1.58 \\
    4 & Xiaomi-Robotics-1 & 20.07/13.93 & 23.54/17.00 & 26.69/18.83 & 38.39/23.67 & 7.81/6.56 & 3.94/3.58 \\
    5 & OpenWAM-$\alpha$ & 17.18/11.92 & 20.71/14.83 & 18.45/9.25 & 34.93/25.33 & 10.41/9.11 & 1.41/1.08 \\
    6 & Meituan-Robotics-0 & 14.95/9.53 & 13.75/8.17 & 16.77/7.75 & 29.61/18.58 & 10.06/8.89 & 4.54/4.25 \\
    7 & Hy-Embodied-0.5-VLA & 13.07/8.80 & 11.78/8.39 & 13.81/8.00 & 25.74/14.92 & 13.37/12.11 & 0.65/0.58 \\
    8 & Spatial Forcing & 12.38/8.04 & 14.12/9.34 & 17.32/10.58 & 23.26/14.58 & 5.43/4.11 & 1.78/1.58 \\
    9 & Pi-05 & 11.41/6.91 & 13.38/8.17 & 12.40/5.50 & 23.54/14.67 & 5.78/4.56 & 1.98/1.67 \\
    10 & InternVLA-A1.5 & 11.15/7.14 & 10.35/6.83 & 15.23/10.17 & 23.80/13.75 & 4.93/3.56 & 1.43/1.42 \\
    11 & StarVLA-PI\_v3 & 10.81/7.51 & 11.22/8.05 & 17.77/12.50 & 18.46/11.00 & 4.59/4.00 & 2.03/2.00 \\
    12 & VLAct & 10.65/7.58 & 9.54/6.28 & 20.57/15.17 & 20.12/13.67 & 0.66/0.56 & 2.37/2.25 \\
    13 & X-VLA & 10.13/6.52 & 10.47/6.78 & 18.32/12.00 & 16.53/9.75 & 4.76/3.56 & 0.55/0.50 \\
    14 & MolmoAct2 & 8.99/5.03 & 8.53/4.67 & 10.92/4.50 & 15.20/7.00 & 5.53/4.56 & 4.75/4.42 \\
    15 & StarVLA-GR00T & 8.25/4.67 & 7.96/5.28 & 9.56/4.67 & 16.18/7.17 & 5.60/4.33 & 1.95/1.92 \\
    16 & StarVLA-OFT & 8.18/4.82 & 7.41/4.78 & 16.08/10.17 & 14.28/6.17 & 1.74/1.67 & 1.41/1.33 \\
    17 & X-WAM & 7.69/3.83 & 7.39/3.33 & 6.72/1.83 & 17.47/9.08 & 6.32/4.67 & 0.57/0.25 \\
    18 & Xiaomi-Robotics-0 & 6.93/4.18 & 7.43/5.55 & 8.42/4.58 & 13.51/6.92 & 5.07/3.67 & 0.22/0.17 \\
    19 & StarVLA & 6.40/3.24 & 3.94/2.33 & 9.90/4.33 & 14.15/6.50 & 3.34/2.44 & 0.67/0.58 \\
    20 & GigaWorld-Policy-0 & 6.20/3.27 & 5.34/2.89 & 6.15/1.83 & 15.51/8.92 & 3.46/2.22 & 0.54/0.50 \\
    21 & GalaxeaVLA (G0) & 5.82/2.96 & 4.54/2.83 & 8.10/3.83 & 12.60/5.58 & 3.17/1.89 & 0.70/0.67 \\
    22 & LingBot-VLA & 5.50/2.96 & 6.71/4.28 & 5.33/1.83 & 10.89/5.25 & 3.82/2.78 & 0.72/0.67 \\
    23 & EventVLA & 4.97/2.81 & 3.95/1.94 & 10.13/5.75 & 5.05/0.83 & 4.92/4.78 & 0.80/0.75 \\
    24 & AHA-WAM & 4.82/2.39 & 5.79/3.27 & 5.86/2.42 & 8.61/2.67 & 2.97/2.78 & 0.88/0.83 \\
    25 & ABot-M0 & 3.67/1.73 & 5.73/3.50 & 5.50/1.75 & 3.96/0.50 & 2.44/2.22 & 0.72/0.67 \\
    26 & Fast-WAM & 3.48/2.03 & 2.33/1.11 & 1.96/0.00 & 9.14/5.17 & 3.55/3.44 & 0.42/0.42 \\
    27 & Pi-0 & 3.48/1.53 & 3.94/2.55 & 3.56/0.75 & 6.19/2.00 & 3.47/2.11 & 0.25/0.25 \\
    28 & \textbf{DeepSeek-Flash\dag} & 2.99/1.92 & 2.42/1.67 & 1.75/0.00 & 2.12/0.00 & 2.17/1.67 & 6.50/6.25 \\
    29 & GROOT-N1.7 & 2.85/1.31 & 2.16/1.22 & 2.54/0.67 & 8.30/3.58 & 1.06/0.89 & 0.18/0.17 \\
    30 & InternVLA-A1 & 2.48/1.08 & 2.86/1.83 & 3.00/0.92 & 4.79/1.17 & 1.58/1.33 & 0.17/0.17 \\
    31 & SmolVLA (Single Task) & 1.82/0.85 & 1.68/1.22 & 2.87/0.33 & 1.22/0.25 & 3.35/2.44 & 0.00/0.00 \\
    32 & LDA-1B & 1.66/0.54 & 0.81/0.11 & 3.33/0.67 & 1.98/0.08 & 2.11/1.78 & 0.08/0.08 \\
    33 & \textbf{GPT-5.5} & 1.13/0.88 & 0.19/0.00 & 0.42/0.00 & 0.51/0.00 & 1.67/1.67 & 2.87/2.75 \\
    34 & ACT (Single Task) & 1.08/0.47 & 0.69/0.56 & 0.85/0.00 & 1.73/0.92 & 2.15/0.89 & 0.00/0.00 \\
    35 & GO-1 & 0.99/0.53 & 1.58/1.22 & 1.45/0.42 & 1.13/0.25 & 0.70/0.67 & 0.08/0.08 \\
    36 & H-RDT & 0.67/0.12 & 0.49/0.22 & 0.41/0.00 & 2.23/0.17 & 0.12/0.11 & 0.08/0.08 \\
    37 & RDT & 0.51/0.13 & 0.55/0.33 & 0.38/0.00 & 1.13/0.00 & 0.49/0.33 & 0.00/0.00 \\
    38 & DM0 & 0.45/0.05 & 0.49/0.06 & 0.61/0.00 & 0.97/0.08 & 0.20/0.11 & 0.00/0.00 \\
    39 & Dexora-1B & 0.38/0.02 & 0.49/0.11 & 0.49/0.00 & 0.82/0.00 & 0.12/0.00 & 0.01/0.00 \\
    40 & A1 & 0.28/0.02 & 0.16/0.00 & 0.09/0.00 & 1.07/0.00 & 0.00/0.00 & 0.08/0.08 \\
    41 & Spirit v1.5 & 0.23/0.14 & 0.80/0.50 & 0.03/0.00 & 0.11/0.00 & 0.22/0.22 & 0.00/0.00 \\
    42 & TinyVLA & 0.22/0.07 & 0.03/0.00 & 0.05/0.00 & 0.67/0.00 & 0.11/0.11 & 0.25/0.25 \\
    43 & OpenVLA-OFT & 0.21/0.02 & 0.04/0.00 & 0.20/0.00 & 0.70/0.00 & 0.00/0.00 & 0.08/0.08 \\
    \bottomrule
  \end{tabular}}
\end{table}

\begin{table}[p]
  \centering
  \caption{Per-task Score on RoboDojo-Sim (process-reward mean $\times$ 100). Public
  columns copied from the official Per-Task board (2026-09-10); only the strongest four
  public policies are shown here. Astra and GPT-5.5:
  1 seed, 50 episodes per task. \dag~DeepSeek-Flash: 1 seed, 10 episodes per task.
  Generalization cells are the mean of Gen-Std and Gen-Rand. Best value per row in bold.}
  \label{tab:per-task}
  \footnotesize
  \setlength{\tabcolsep}{3pt}
  \resizebox{\linewidth}{!}{%
  \begin{tabular}{lrrrrrrr}
    \toprule
    Task & GPT-6 Astra & GPT-5.5 & DS-Flash\,\dag & DM0.5 & Galaxea G0.5 & Xiaomi-1 & OpenWAM-$\alpha$ \\
    \midrule
    \multicolumn{8}{l}{\textit{Generalization}} \\
    \texttt{stack\_bowls} & \textbf{66.5} & 0.3 & 3.0 & 30.5 & 42.2 & 56.4 & 49.9 \\
    \texttt{push\_T} & \textbf{60.0} & 0.0 & 0.0 & 0.0 & 0.0 & 0.7 & 0.0 \\
    \texttt{pack\_objects\_into\_box} & 11.6 & 0.2 & 2.0 & 15.1 & 18.2 & 19.9 & \textbf{22.5} \\
    \texttt{fold\_clothes} & \textbf{76.4} & 1.2 & 2.0 & 32.5 & 36.3 & 47.3 & 57.5 \\
    \texttt{hang\_mugs} & 0.9 & 0.0 & 0.0 & 9.0 & 8.6 & 15.7 & \textbf{16.3} \\
    \texttt{sweep\_blocks} & \textbf{4.0} & 0.0 & 0.0 & 4.0 & 2.0 & 2.7 & 1.3 \\
    \texttt{pour\_liquid\_into\_cup} & 4.0 & 0.0 & 0.0 & 34.7 & 35.3 & \textbf{42.0} & 34.7 \\
    \texttt{make\_toast} & 0.5 & 0.0 & 0.0 & 4.5 & 12.0 & \textbf{12.7} & 12.5 \\
    \texttt{arrange\_largest\_number} & \textbf{68.7} & 0.0 & 0.5 & 9.6 & 4.7 & 10.4 & 4.9 \\
    \texttt{sort\_nesting\_dolls\_by\_size} & \textbf{18.0} & 0.0 & 0.0 & 10.7 & 7.3 & 13.3 & 4.7 \\
    \texttt{store\_laptop\_and\_headphones} & 1.6 & 0.0 & 0.0 & 16.9 & 28.3 & \textbf{44.4} & 30.5 \\
    \texttt{stack\_blocks} & \textbf{88.1} & 0.6 & 21.5 & 21.8 & 26.5 & 17.1 & 13.7 \\
    \addlinespace
    \multicolumn{8}{l}{\textit{Precision}} \\
    \texttt{fasten\_screws} & 36.4 & 0.0 & 2.0 & \textbf{40.2} & 30.0 & 22.9 & 24.7 \\
    \texttt{insert\_tubes} & 6.0 & 0.0 & 0.0 & \textbf{71.7} & 58.5 & 50.0 & 26.3 \\
    \texttt{plug\_in\_charger} & 0.0 & 0.0 & 0.0 & 2.7 & 0.7 & \textbf{4.0} & 4.0 \\
    \texttt{pour\_balls\_into\_vase} & 4.0 & 0.0 & 0.0 & 9.3 & 28.0 & \textbf{46.0} & 17.3 \\
    \texttt{play\_Xylophone} & \textbf{8.0} & 0.0 & 0.0 & 0.7 & 0.0 & 0.0 & 0.0 \\
    \texttt{deposit\_coin} & 16.0 & 1.6 & 4.0 & 13.9 & 10.9 & \textbf{24.7} & 9.1 \\
    \texttt{insert\_key} & 14.4 & 1.8 & 6.0 & 4.9 & \textbf{14.9} & 13.3 & 13.7 \\
    \texttt{build\_tower} & 16.4 & 0.0 & 2.0 & 55.2 & \textbf{82.9} & 52.6 & 52.5 \\
    \addlinespace
    \multicolumn{8}{l}{\textit{Long-Horizon}} \\
    \texttt{put\_bottles\_into\_dustbin} & 35.3 & 1.3 & 2.0 & 81.7 & 96.3 & \textbf{97.7} & 94.0 \\
    \texttt{play\_tic\_tac\_toe} & 10.7 & 0.8 & 5.0 & 35.8 & \textbf{65.2} & 36.8 & 64.1 \\
    \texttt{classify\_objects} & \textbf{69.0} & 0.3 & 3.0 & 26.8 & 10.3 & 17.2 & 5.5 \\
    \texttt{fill\_pen\_holder} & 5.0 & 0.0 & 0.0 & 22.1 & 41.3 & \textbf{44.7} & 15.4 \\
    \texttt{fill\_egg\_holder} & 5.5 & 0.2 & 2.0 & 1.4 & 3.0 & \textbf{11.7} & 5.8 \\
    \texttt{organize\_table} & 39.5 & 1.5 & 5.0 & 44.0 & 46.3 & 57.7 & \textbf{62.5} \\
    \texttt{play\_stacking\_toy} & \textbf{6.6} & 0.0 & 0.0 & 1.2 & 0.5 & 0.0 & 0.1 \\
    \texttt{make\_kong} & 0.0 & 0.0 & 0.0 & 56.7 & \textbf{90.0} & 41.3 & 32.0 \\
    \addlinespace
    \multicolumn{8}{l}{\textit{Memory}} \\
    \texttt{cover\_blocks} & 49.3 & 0.0 & 3.0 & \textbf{100.0} & 20.7 & 17.7 & 22.2 \\
    \texttt{match\_and\_pick\_from\_conveyor} & 62.0 & 10.0 & 10.0 & \textbf{70.7} & 29.3 & 26.7 & 36.7 \\
    \texttt{swap\_T} & \textbf{18.0} & 0.0 & 0.0 & 0.0 & 0.0 & 0.0 & 0.0 \\
    \texttt{press\_by\_number} & 70.0 & 0.0 & 0.0 & \textbf{95.3} & 0.0 & 0.0 & 0.0 \\
    \texttt{imitate\_sorting\_sequence} & \textbf{58.9} & 0.0 & 0.0 & 1.8 & 1.7 & 2.5 & 2.9 \\
    \texttt{swap\_blocks} & 0.0 & 0.0 & 0.0 & \textbf{18.7} & 0.0 & 0.0 & 0.7 \\
    \addlinespace
    \multicolumn{8}{l}{\textit{Open}} \\
    \texttt{align\_blocks} & \textbf{50.0} & 0.0 & 20.0 & 0.0 & 0.0 & 0.0 & 0.0 \\
    \texttt{solve\_equation} & \textbf{40.0} & 2.0 & 0.0 & 0.0 & 0.0 & 0.0 & 0.0 \\
    \texttt{stack\_blocks\_by\_language} & \textbf{48.0} & 0.8 & 2.0 & 4.8 & 0.1 & 1.3 & 0.9 \\
    \texttt{general\_pickup} & \textbf{84.0} & 20.0 & 30.0 & 14.0 & 12.7 & 28.0 & 8.7 \\
    \texttt{classify\_objects\_by\_language} & \textbf{46.0} & 0.2 & 0.0 & 0.5 & 1.1 & 2.0 & 1.3 \\
    \texttt{pick\_from\_conveyor\_by\_image} & \textbf{4.0} & 0.0 & 0.0 & 0.0 & 0.0 & 0.0 & 0.0 \\
    \texttt{store\_tools\_in\_toolbox} & \textbf{2.5} & 0.0 & 0.0 & 0.2 & 0.0 & 0.2 & 0.3 \\
    \texttt{pour\_by\_language} & \textbf{0.4} & 0.0 & 0.0 & 0.0 & 0.0 & 0.0 & 0.0 \\
    \addlinespace
    \bottomrule
  \end{tabular}}
\end{table}

\section{Details of one-shot in-context learning}
\label{sec:app-oneshot}

This appendix records the protocol, condition definitions, and task-level
evidence for the one-shot demonstration-conditioned ICL experiment in
Section~\ref{sec:icl-demo} and Table~\ref{tab:f3}.

\subsection{Protocol}
\label{sec:app-oneshot-protocol}

Every run uses the interface of Section~\ref{sec:study-design}: the same
\texttt{move\_eef} and \texttt{give\_up} tools, the same bounding and
interpolation of an accepted target, and the same observation contract. The
demonstration is the intended difference from the zero-shot prompt.

The design is matched. Each condition covers the same 34 tasks and the same ten
layouts per task, so each condition is 340 task--layout pairs at one episode per
pair on a single seed. The two demonstration conditions are 340 new episodes
each. The zero-shot column is not a new run: it is the official 50-episode
Astra campaign restricted to those exact 340 task--layout pairs. Because the
pairs are identical across conditions, the zero-shot column is the control for
the other two, and a gap between columns cannot come from a difference in task or
layout difficulty.

The zero-shot cell of Table~\ref{tab:f3} is a micro success rate over these 340
pairs, $78/340 = 22.9\%$. It is close to the $472/2100 = 22.48\%$ micro rate of
the official board (Section~\ref{sec:results-board}), but it is computed over a
different task set and a different number of episodes per task, so only the
within-table comparison carries weight.

\subsection{What each condition puts in the prompt}
\label{sec:app-oneshot-conditions}

The demonstration always comes from a \emph{different layout of the same task}.
It is therefore a correct solution to the task the model faces, recorded in a
different scene. The experiment tests transfer from that example, not replay of
a solution for the current layout.

The image condition supplies a sequence of camera images together with the
end-effector pose recorded at each of them. The poses are in the same
dimensions \texttt{move\_eef} takes, so the model is shown, in its own action
vocabulary, a sequence of targets that solved the task once. The text condition
describes the same trajectory in prose and supplies no images. Neither condition
provides a demonstration recorded in the current layout.

\subsection{Task-level evidence}
\label{sec:app-oneshot-subsets}

\begin{table}[htbp]
  \centering
  \caption{Where the demonstration conditions lose episodes. The 34 tasks split
  into those the zero-shot controller never solves and those it solves at least
  once. The first row is the subset count reported in
  Section~\ref{sec:icl-demo}; the second row is the total of
  Table~\ref{tab:f3} minus the first row. No cell is a new measurement.}
  \label{tab:a4}
  \small
  \begin{tabular}{llrrr}
    \toprule
    Task subset & Pairs & Zero-shot & Image demo & Text demo \\
    \midrule
    Zero-shot 0/10 (15 tasks)          & 150 & 0/150 = 0.0\%   & 2/150 = 1.3\%   & 0/150 = 0.0\% \\
    At least one success (19 tasks)     & 190 & 78/190 = 41.1\% & 59/190 = 31.1\% & 44/190 = 23.2\% \\
    \midrule
    All tasks (34 tasks)               & 340 & 78/340 = 22.9\% & 61/340 = 17.9\% & 44/340 = 12.9\% \\
    \bottomrule
  \end{tabular}
\end{table}

Table~\ref{tab:a4} splits the 34 tasks by what the zero-shot controller already
does with them. Fifteen tasks are 0/10 zero-shot; the reported subset counts
cover 150 pairs per condition, which is those fifteen tasks at ten layouts each.
The remaining nineteen tasks hold every zero-shot success, since the first
subset contributes none by construction.

On the fifteen tasks with no matched zero-shot success, the image condition
succeeds in 2 of 150 episodes and the text condition in none. On the remaining
nineteen tasks, success drops from 41.1\% to 31.1\% with images and to 23.2\%
with text. These are net losses of 19 and 34 successes within that subset.
The two additional image-condition successes in the first subset reduce its
overall deficit to 17 episodes. These counts describe outcomes under this
protocol, not whether a task is fundamentally beyond the model's capability.

\paragraph{Paired outcome changes.} Against the
zero-shot column, an image demonstration turns 31 failures into successes and 48
successes into failures. A text demonstration turns 18 up and 52 down. A
demonstration therefore changes 79 of the 340 pairs in the image condition and 70
in the text condition. More pairs change from success to failure than in the
opposite direction. The net losses above are the difference between the two
directions, not the number of pairs a demonstration changes. The losses all fall
on the nineteen tasks with at least one zero-shot success, since the other fifteen
have no success to lose: 29 of the 31 image-condition gains and all 48 of its
losses sit in that subset. Counted by task instead of by episode, images help 7 of
the 34 tasks and hurt 12, and text helps one.

\subsection{Possible explanations for the decline}
\label{sec:app-oneshot-failures}

The experiment measures the effect of adding these demonstrations, but does not
isolate why aggregate success decreases. When the model already interprets an
instruction correctly, another example may not resolve an execution-precision
limitation. Transferring layout-specific contact geometry may also lead to errors.
Neither explanation is established by the aggregate results. Demonstration
format, context length, and stochastic variation remain alternative explanations.

\begin{table}[htbp]
  \centering
  \caption{An illustrative task from the demonstration campaign. Each condition
  contains ten matched task--layout pairs. Higher success on this task does not
  identify the cause of the difference.}
  \label{tab:a5}
  \small
  \begin{tabular}{llll}
    \toprule
    Task & Zero-shot & With demonstration & Condition \\
    \midrule
    \texttt{press\_by\_number}           & 5/10 & 7/10 & image \\
    \texttt{press\_by\_number}           & 5/10 & 8/10 & text \\
    \bottomrule
  \end{tabular}
\end{table}

Table~\ref{tab:a5} illustrates a task with higher success under both demonstration
conditions, \texttt{press\_by\_number}. Its 5/10, 7/10, and 8/10 counts do not
establish whether the difference comes from task interpretation or execution.
The 70.0 Score in Table~\ref{tab:per-task} is a different metric from the full
50-episode campaign and should not be compared directly with these counts.

\subsection{What this design does not control}
\label{sec:app-oneshot-limits}

Every cell is one episode per task--layout pair on a single seed. Small per-task
differences may reflect stochastic variation; even the larger subset totals do
not identify the cause of the decline. The experiment covers the one-demonstration setting only, so
it does not establish what several demonstrations, or a demonstration from the
current layout, would do. The 34 tasks are a subset of the official 42, so the
22.9\% baseline should not be read as a restatement of the official Average.
Table~\ref{tab:a5} is illustrative rather than a full task-level breakdown.

\section{Perturbation protocol and episode traces}
\label{sec:app-perturbation}
\label{sec:app-self-improve}

This appendix records the protocol, the perturbation definitions, and the
episode-level evidence behind Section~\ref{sec:icl-online} and
Table~\ref{tab:f4}.

\subsection{Protocol}
\label{sec:app-perturbation-protocol}

Every run uses the \texttt{move\_eef} and \texttt{give\_up} loop of
Section~\ref{sec:study-design}, with the condition-specific changes described below. The task
is \texttt{general\_pickup}. The layouts are 0, 1, 2, 4, 7, 8, 9, and 10, with
one episode per layout on a single seed. The context policy is
$\texttt{image\_horizon} = 2$, so the two most recent observations keep their
images while all earlier text remains. The budget is 100 model calls per episode
unless a row states otherwise, and RoboDojo's own step limit runs down in
parallel.

These eight layouts are the ones the unperturbed controller solves. Unperturbed,
the eight episodes are 8/8 at process score 1.00, against 42/50 for the same task
under the official 50-episode protocol. The baseline row therefore sits at the
ceiling by construction, and a perturbation can only move the count down.

The perturbation is active from the first turn. No explicit explanation of the
perturbation is added to the system or Goal message. However, the
\texttt{negate\_xyz} tool description exposes transformed coordinate bounds:
for example, the advertised z range is $[-1.565,-0.7]$ rather than
$[0.7,1.565]$. The model-visible interface is therefore not identical across
conditions, and these bounds may provide a cue to the transformation.
Model weights remain fixed and no memory is carried between episodes.
Table~\ref{tab:f4} includes no retries after failure. A separate larger-budget
rerun is reported below and is excluded from that table.

\subsection{What each perturbation changes}
\label{sec:app-perturbation-probes}

Four probes change the visual observation. \emph{Top--bottom image flip}
(\texttt{flip\_vision\_ud}) flips every camera image before the model receives it.
\emph{Left--right image mirror} (\texttt{flip\_vision\_lr}) mirrors every camera
image, reversing apparent left--right positions without changing camera identities. \emph{No
head-camera view} masks \texttt{cam\_head} and leaves the two wrist views.
\emph{Right-wrist view only} masks the head and left-wrist cameras.

One probe changes the coordinate mapping. \emph{Negated Cartesian axes}
(\texttt{negate\_xyz}) uses sign-reversed Cartesian target coordinates, with
correspondingly transformed bounds in the tool description. This changes the
coordinate convention rather than simply reversing a displacement from the
current pose. One probe changes action execution.
\emph{Per-move pose jitter} adds a random offset to every commanded pose. The
displacement has a mean magnitude of 10\,cm and is resampled on each move.

The model receives state text, action feedback, and the available camera views.
These traces do not isolate the contribution of each feedback channel to recovery.

\subsection{The conditions without a figure}
\label{sec:app-perturbation-remaining}

Figure~\ref{fig:icl-perturbation} in Section~\ref{sec:icl-online} shows Left--right
image mirror, Right-wrist view only, and Per-move pose jitter. The three remaining
perturbation rows of Table~\ref{tab:f4} are recorded here.

All eight layouts succeed under Top--bottom image flip. Six succeed without
the head-camera view, compared with three under Right-wrist view only. Selected
camera-masking failures include prolonged visual search
(Appendix~\ref{sec:app-perturbation-failures}); the counts alone do not establish
why the conditions differ.

Four layouts succeed under Negated Cartesian axes. In the layout 0 trace,
the model revises its forward/back estimate after observing the wrist view.
Appendix~\ref{sec:app-perturbation-verbatim} reproduces selected calls. This is
evidence of a spatial correction, not proof that the model identified the full
coordinate transformation.

\subsection{Episode-level evidence}
\label{sec:app-perturbation-traces}

\begin{table}[htbp]
  \centering
  \caption{Selected episodes from the perturbation sweep. Calls counts the model
  calls spent; a failed episode reports the calls used before the budget or the
  episode ended. All rows are \texttt{general\_pickup} at a 100-call budget except
  the last, which repeats layout 9 at 170 calls and 400 environment steps.}
  \label{tab:a3}
  \small
  \begin{tabular}{llll}
    \toprule
    Condition & Layout & Outcome & Calls \\
    \midrule
    Top--bottom image flip       & 0 & success & 14 \\
    Left--right image mirror     & 0 & success & 15 \\
    Negated Cartesian axes       & 0 & success & 19 \\
    Negated Cartesian axes       & 4 & failure & 22 of 100 \\
    No head-camera view          & 0 & success & 16 \\
    Per-move pose jitter         & 0 & success & 14 \\
    Per-move pose jitter         & 4 & failure & 18 of 100 \\
    Right-wrist view only        & 9 & failure & 25 of 100 \\
    Right-wrist view only, larger budget & 9 & success & 38 \\
    \bottomrule
  \end{tabular}
\end{table}

Table~\ref{tab:a3} lists selected episodes from the sweep. The
diagnoses are in the model's own notes. Under Negated Cartesian axes on layout 0,
the fifth call states that its forward/back estimate was reversed; that episode
is excerpted in Table~\ref{tab:a6}. Under Per-move pose jitter, the second call
reads: ``The rotation reached 45
degrees, but the position missed by 129 mm.'' Under Left--right image mirror
(\texttt{flip\_vision\_lr}), the correction is an arm reassignment: ``The camera
views show that the right arm
is the one beside the scissors. I am restoring the idle left arm and preparing
the right hand for a downward grasp.'' That note is the one quoted in
Figure~\ref{fig:icl-perturbation}, between the two turns shown there. The notes
refer to both visual observations and motion feedback; they do not establish a
text-only recovery mechanism.

Right-wrist view only includes a supplementary example of one arm observing
while the other grasps. Layout 9 fails at the standard budget after spending fifteen calls
rejecting a quadcopter, a wrench, and a tape measure before locating the
corkscrew. Re-run at 170 calls and 400 environment steps, the same layout
succeeds in 38 calls: the right arm sweeps, tilts, and rolls in order to look
(``Only the right wrist camera is available and the corkscrew is not yet
visible''), and the other hand performs the grasp (``I bring the left hand above
the handle while the right wrist watches the approach''). One arm becomes a
camera and the other a manipulator. This follows general advice in the system
prompt to use an idle wrist camera when the working view is insufficient. It
does not establish that the model discovered the role split without guidance.

\subsection{Success and observable correction}
\label{sec:app-perturbation-modes}

An episode that succeeds without a visible change in behaviour may reflect
existing robustness rather than adaptation. Other episodes include notes about
an unexpected outcome followed by a change in actions. These provide evidence
of observable correction, but the notes alone do not establish the internal
mechanism or distinguish adaptation from ordinary closed-loop control.
Table~\ref{tab:f4} reports final outcomes; Table~\ref{tab:a6} illustrates a
sequence of corrections. Neither measures the causal benefit of retaining history.

\subsection{Selected calls from a successful episode}
\label{sec:app-perturbation-verbatim}

\begin{table}[htbp]
  \centering
  \caption{Selected verbatim notes from a successful \texttt{negate\_xyz} episode
  on \texttt{general\_pickup} layout 0, 19 calls, success. The note is the text
  the model attached to its own \texttt{move\_eef} call, followed by the
  dimensions that call requested. Call 5 refers to the wrist view and revises the
  forward/back estimate; its requested motion is rejected as unreachable.
  Targets use the transformed bounds described in Appendix~\ref{sec:app-perturbation-probes}.}
  \label{tab:a6}
  \small
  \begin{tabular}{@{}r p{0.88\linewidth}@{}}
    \toprule
    Call & Model's own note, and the targets it requested \\
    \midrule
    4  & ``The scissors are ahead and slightly to the right in the wrist view. I
         will move toward their hinge while staying above the tabletop.''
         \newline
         \texttt{right\_x} $-0.40$, \texttt{right\_y} $0.38$,
         \texttt{right\_z} $-1.07$ \\
    \addlinespace
    5  & ``The wrist view now shows the near table edge, so the forward/back
         estimate was reversed. I will move back across the table toward the
         scissors.''
         \newline
         \texttt{right\_y} $0.02$, \texttt{right\_z} $-1.18$ \\
    \addlinespace
    12 & ``The closer view places the hinge slightly right and ahead of the jaw
         center. I will make a small lateral correction over the metal hinge.''
         \newline
         \texttt{right\_x} $-0.389$, \texttt{right\_y} $0.025$ \\
    \addlinespace
    \multicolumn{2}{@{}l}{\textit{Intervening calls omitted.}} \\
    \addlinespace
    17 & ``The gripper has closed against the scissors rather than fully
         shutting. I will lift a few millimetres first to confirm they are
         secure.''
         \newline
         \texttt{right\_z} $-0.78$ \\
    \addlinespace
    19 & ``The scissors are securely held and clear of the table. I will lift
         them to more than 10 cm above their starting height.''
         \newline
         \texttt{right\_z} $-0.89$ \\
    \bottomrule
  \end{tabular}
\end{table}

Table~\ref{tab:a6} reproduces selected notes and requested targets from the
\texttt{negate\_xyz} layout 0 episode. After call 4, call 5 refers to the wrist
view and revises the forward/back estimate. That call is rejected as unreachable;
later calls make further corrections and the episode eventually succeeds.
The displayed targets use the transformed coordinate bounds, not the normal
bounds in Appendix~\ref{sec:app-prompt}. The trace does not establish a one-call
adaptation cost or equivalence to an unperturbed trajectory.

The same episode also shows the limit of this evidence. The recovery is visible
because the model wrote it down. Nothing in the trace establishes that the model
could not have recovered without stating the diagnosis, and nothing establishes
that a pretrained policy could not recover in some other way.

\subsection{Failure modes}
\label{sec:app-perturbation-failures}

Selected failures include extended visual search under camera masking and
repeated grasp corrections under pose jitter. The Negated Cartesian axes run on
layout 4 ends after 22 calls without success. Under Right-wrist view only on
layout 0, the model reports that the object slipped during lifting and then
attempts further grasps. These observations suggest multiple failure modes, but
do not isolate their causes or account for every failed layout.

\paragraph{An excluded run.} On layout 4 the prompt stated that a lift may be
shared between both hands, and the controller produced a two-handed pickup in 23
calls. It opened both grippers at opposite ends of a toy car, rotated both jaws
by 75 degrees, seated each hand over a separate axle, and lifted in synchronized
5\,mm steps. That run adds explicit task guidance rather than only perturbing the interface,
so it is not a row of Table~\ref{tab:f4}.

\subsection{What this design does not control}
\label{sec:app-perturbation-limits}

Each cell of Table~\ref{tab:f4} contains only eight binary trials on one seed,
so small differences should not be treated as reliable condition rankings.
The layouts are matched across conditions but selected for unperturbed success,
which limits generalization to other layouts. There is no history ablation or
pretrained-policy comparison under these perturbations. Thus the probe does not
establish that correction depends on retained history or is distinctive of LLMs.
The transformed tool bounds also limit claims about discovering an entirely
unannounced coordinate mapping.

\section{Real-robot diagnostic material}
\label{sec:app-real}

These tables preserve the detailed material behind Section~\ref{sec:real}. The
official RoboDojo-Real protocol was not completed with GPT-6 Astra. The retained
samples and separate joint-position deployments are diagnostic material, not
official results.
Table~\ref{tab:real} reports the 12-task, 33-trial material retained around the
safety stop, with variable $n = 1$--$4$ per task across ARX X5 (9 clips), Piper
(21), and Piper X (3); trial scores come directly from each task's
\texttt{scores.json}. \texttt{stack\_bowls} \texttt{trial\_8} is the only full
success; its trials 4, 5, and 6 and \texttt{store\_in\_safe} \texttt{trial\_1}
receive partial credit, and zero-score clips often show approach, grasp, or
transport attempts with the terminal predicate unmet. The official
complete-protocol rows are context only and are not directly comparable.

Table~\ref{tab:r1} collects the qualitative notes from the joint-position
deployments on the Franka and the two-arm humanoid. These use a different
interface from the evaluated \texttt{move\_eef} controller and carry no
comparable Score or SR. The notes do not establish a controlled comparison
between the embodiments or identify a cause for differences in behaviour.
Images of the humanoid deployment are not included.
The retained operator notes also record that the model did not adopt large
rotations to improve contact on either body and sometimes described dynamic
corrections too late to execute them.

\begin{table}[htbp]
  \centering
  \caption{RoboDojo-Real diagnostic sample. Diagnostic rows report Astra's retained 12-task, 33-trial material; Score is the mean process score $\times$ 100 and SR is the full-success rate. The official 18-task rows are complete-protocol references and are not directly comparable.}
  \label{tab:real}
  \small
  \begin{tabular}{llrlrr}
    \toprule
    Type & Entry & $n$ & Raw trial scores & Score & SR \\
    \midrule
    ARX X5 & \texttt{cover\_blocks} & 2 & 0 / 0 & 0 & 0\% \\
    ARX X5 & \texttt{insert\_tubes} & 3 & 0 / 0 / 0 & 0 & 0\% \\
    ARX X5 & \texttt{make\_bread} & 1 & 0 & 0 & 0\% \\
    ARX X5 & \texttt{make\_food} & 2 & 0 / 0 & 0 & 0\% \\
    ARX X5 & \texttt{store\_in\_safe} & 1 & 0.4 & 40 & 0\% \\
    Piper & \texttt{fill\_pen\_holder} & 3 & 0 / 0 / 0 & 0 & 0\% \\
    Piper & \texttt{insert\_charger} & 3 & 0 / 0 / 0 & 0 & 0\% \\
    Piper & \texttt{put\_objects\_into\_basket} & 3 & 0 / 0 / 0 & 0 & 0\% \\
    Piper & \texttt{stack\_and\_cover\_blocks} & 4 & 0 / 0 / 0 / 0 & 0 & 0\% \\
    Piper & \texttt{stack\_bowls} & 4 & 0.3 / 0.3 / 0.3 / 1.0 & 47.5 & 25\% \\
    Piper & \texttt{stand\_up\_bottles} & 4 & 0 / 0 / 0 / 0 & 0 & 0\% \\
    Piper X & \texttt{sweep\_blocks} & 3 & 0 / 0 / 0 & 0 & 0\% \\
    \addlinespace
    Diagnostic & Pooled retained clips & 33 & --- & 6.97 & 3.03\% \\
    \addlinespace
    Official 18-task & OpenWAM-$\alpha$ & 18 tasks & --- & 37.60 & 24.40\% \\
    Official 18-task & Pi-05 & 18 tasks & --- & 22.90 & 12.80\% \\
    Official 18-task & InternVLA-A1 & 18 tasks & --- & 12.00 & 7.20\% \\
    \bottomrule
  \end{tabular}
\end{table}

\begin{table}[htbp]
  \centering
  \caption{Qualitative notes from separate joint-position deployments, not an
  official Score/SR. Franka clips are shown in the online report's hardware section;
  the humanoid rows have no accompanying media. The report is available at
  \url{https://robodojo-benchmark.com/report/gpt-6-astra-eval}.}
  \label{tab:r1}
  \small
  \begin{tabular}{llp{0.36\linewidth}l}
    \toprule
    Body & Task & Observation & Media \\
    \midrule
    Franka & Pour water & initial miss; later flow reaches cup, then stops & Report \\
    Franka & Cup on shelf & placement and mid-air drop observed & Report \\
    Humanoid & Pick and place & rigid and soft objects manipulated & Not available \\
    Humanoid & Make breakfast & toast-into-slot fails & Not available \\
    Humanoid & Pour water & high pour, some spill & Not available \\
    Humanoid & Fold a shirt & corners first; neatness stalls & Not available \\
    Humanoid & Draw & color OK; later uses idle wrist cam & Not available \\
    \bottomrule
  \end{tabular}
\end{table}

\section{Mobile humanoid demonstrations}
\label{sec:app-demos}

These two recordings show a mobile humanoid performing walking and grasping
sequences commanded by Astra through a separate interface. They are qualitative
demonstrations, with no success rate or process score. The action vocabulary
includes locomotion commands rather than only joint-position targets. The
recordings alone do not establish how each command is implemented by the
low-level controller, so we do not classify them under the strict definition
in Section~\ref{sec:study-boundary}. Section~\ref{sec:beyond} introduces the
recordings; this appendix records their contents.

\paragraph{Pickup.} The model walks to a target object and grasps it. The action
vocabulary is \texttt{Walk}, \texttt{turn}, \texttt{hand cm}, \texttt{wrist R/P/Y},
and \texttt{close}; the overlay in the recording is the request. The task is
\texttt{target RED}, actions 004--026, with \texttt{close 1.00} at action 022. The
recording runs 23\,s across five views: two external, plus ego, left wrist, and
right wrist. The left hand performs the grasp.

\paragraph{Long-horizon sit-and-pick.} The model walks, sits down, and grasps while
seated in a longer recorded sequence. The task is
\texttt{target WHITE}, 117\,s, actions 001--038, with \texttt{close 1.00} at action
037.

\section{When the model writes the controller instead of acting}
\label{sec:app-piano}

In the piano runs, GPT-6 Astra writes a controller rather than selecting targets
turn by turn: a \texttt{play.py} that
emits 45-D joint targets every 0.05\,s for two Shadow hands on an 88-key piano in
RoboPianist~\citep{robopianist}. The LLM weights remain fixed, but the model refines
the program through simulation trials without access to a reference solution. This is
outside the LLM-as-policy condition of Section~\ref{sec:study-design} and outside the
RoboDojo protocol. It examines code-based controller design rather than
turn-by-turn action selection.
Section~\ref{sec:beyond} introduces the evaluation; this appendix keeps the
isolation rules, runs, and transfer tests.

The controller is written by GPT-6 Astra, driven through the Codex agent scaffold
at reasoning effort \texttt{high}; the model under test is the same one as in the
main experiments, and Codex is only the shell that runs it. There are no motion
primitives: every finger, wrist, and slide target is code the model wrote. The
two-hand Twinkle RL reference scores F1 0.8863 when replayed through our evaluation
setup. We use this as a reference point for the recorded piano runs, not as a
comparison of training efficiency. F1 and precision/recall (P/R) below denote the
piano evaluator's reported metrics, rather than a RoboDojo Score or success rate.

\paragraph{What the model may read.} The score, meaning pitch, onset and offset,
the left and right staff split, and pedal marks; the position and size of all 88
keys; its own joint angles and a forward-kinematics call; and after each attempt,
which keys sounded when, the resulting F1, and rendered frames.

\paragraph{What is withheld.} Human fingering annotations, which ship with the
RoboPianist data and were removed; the RL reference actions and code; the
simulator source and model files; contact forces and other internal state; network
access; and any memory of an earlier session. The protocol excludes fitting a
neural policy, reinforcement learning, and behaviour cloning. Code revision,
search, optimization, and trial and error are permitted. Thus the runs are not
practice-free, even though the LLM weights remain fixed.

\paragraph{Isolation and audit.} The simulator runs in a separate process behind a
socket, the rules are stated in the task file the model reads, and every shell
command and tool call is audited afterwards. Across the three runs, the audit
recorded 201 shell commands, six image views, and zero network calls; it flagged
no violations. The host had user namespaces disabled, so no system-level sandbox
was available. Access restrictions therefore relied on the stated rules and audit,
not on enforced system-level isolation.

\begin{table}[htbp]
  \centering
  \caption{Piano-performance F1 from the final verification episode of each run.
  Tries counts practice episodes plus that verification. Two-hand Twinkle reaches
  0.902; the RL reference replay scores 0.886 in this setup.}
  \label{tab:a2}
  \small
  \begin{tabular}{lrrlrr}
    \toprule
    Score & Notes & F1 & P / R & Tries & Wall \\
    \midrule
    Twinkle, one hand & 14 & 0.907 & 1.00 / 0.91 & 22 & 31 min \\
    Twinkle, two hands & 34 & 0.902 & 0.99 / 0.89 & 33 & 34 min \\
    Chopin nocturne excerpt & 117 & 0.599 & 0.85 / 0.58 & 150 & 40 min \\
    \bottomrule
  \end{tabular}
\end{table}

Table~\ref{tab:a2} reports both practice and final verification. One-hand Twinkle
reaches F1 0.907 after 22 total episodes, with precision 1.00. Two-hand Twinkle
reaches a similar F1 of 0.902 after 33 total episodes; all 35 sounded notes are
assigned to the hand specified by the score. The 117-note nocturne excerpt
reaches F1 0.599 after 150 total episodes, with precision 0.85 and recall 0.58.
The model stopped below its target of 0.9. Lower recall is consistent with
missed or insufficiently sustained notes, but the metrics alone do not isolate
reachability, timing, and coordination as causes.

\subsection{Does the delivered program survive a change of music?}
\label{sec:app-piano-generalization}

We also test the delivered controller on different music. We re-ran
the two-hand Twinkle \texttt{play.py} with no re-optimization on transposed and
time-stretched versions of its own score, and on three scores it had never seen.
The program, the 0.05\,s control loop, and the piano evaluation metric are
identical across rows.

\begin{table}[htbp]
  \centering
  \caption{The delivered two-hand Twinkle program re-run without modification.
  Notes is the score length; Correct hand counts sounded notes assigned to the
  staff the score specifies. Time factors multiply note timestamps, not tempo.
  P/R and F1 are reported by the piano evaluator. One episode per row.}
  \label{tab:a1}
  \small
  \begin{tabular}{llrrlr}
    \toprule
    Score & Change & Notes & F1 & P / R & Correct hand \\
    \midrule
    Twinkle, two hands & as delivered & 34 & 0.902 & 0.99 / 0.89 & 35/35 \\
    Twinkle, two hands & time $\times$0.85 & 34 & 0.883 & 0.98 / 0.87 & 32/32 \\
    Twinkle, two hands & $-$7 semitones, time $\times$1.25 & 34 & 0.837 & 0.94 / 0.82 & 33/38 \\
    Twinkle, two hands & +5 semitones, time $\times$1.1 & 34 & 0.805 & 0.93 / 0.79 & 30/33 \\
    Twinkle, two hands & +2 semitones & 34 & 0.683 & 0.89 / 0.65 & 32/36 \\
    Twinkle, two hands & $-$3 semitones & 34 & 0.655 & 0.85 / 0.63 & 31/37 \\
    C major scale & unseen score & 30 & 0.903 & 0.95 / 0.93 & 30/30 \\
    C major chords & unseen score & 16 & 0.654 & 0.98 / 0.56 & 12/12 \\
    D major scale & unseen score & 30 & 0.556 & 0.81 / 0.56 & 23/34 \\
    \bottomrule
  \end{tabular}
\end{table}

Transfer varies across the tested changes. Reducing note times by a factor of
0.85, equivalent to a 17.6\% faster tempo, lowers F1 by 0.019. All sounded notes
remain assigned to the correct hand. Pure transpositions of $+2$ and $-3$
semitones lower F1 by approximately 0.22--0.25. The two transposed rows above
0.80 also change timing, so they do not isolate transposition.
On unseen music, the C major scale reaches 0.903 with 30 of 30 sounded notes
assigned to the correct hand. The D major scale reaches 0.556, with 23 of 34
sounded notes assigned correctly. For C major chords, precision is 0.98 but only
12 notes sound for a 16-note score. These outcomes suggest limitations in transfer,
but do not isolate black-key geometry, timing, or simultaneous pressing as causes.

Beyond writing controllers, the model also produced the performance itself as a
fixed sequence of joint targets, refining each trajectory through repeated
simulation trials and submitting 161 keyframes at 50\,ms intervals. Open-loop
replay achieved F1 scores of 0.901 on one-hand Twinkle and 0.920 on two-hand
Twinkle.

\section{Limitations}
\label{sec:limitations}

This appendix states the limits of the evaluation. Observed model failure modes
are described in Section~\ref{sec:results-wall}.
\\[2pt]
\textbf{The detailed analysis concerns one model.} The qualitative capability
analysis and Section~\ref{sec:icl} focus on Astra. The benchmark and ICL probes
use one evaluation seed, without repeated trials to estimate variability.
Three models cannot establish a general capability threshold for LLMs.
DeepSeek-Flash is also evaluated for 10 episodes per task rather
than 50, and the eight perturbation layouts were selected as those the unperturbed
model solves, which puts that baseline at the ceiling by construction and leaves
Table~\ref{tab:f4} too small for a reliable quantitative comparison.
\\[2pt]
\textbf{The action and observation spaces define a measurement boundary.}
\texttt{move\_eef} is the only motion tool. A task requiring continuous contact or
non-quasi-static motion may fail because this action representation cannot express the
behaviour, rather than because the model cannot conceive it. Visual observations
are RGB-only: no metric depth is exposed, so part of the precision ceiling may come from
the inputs rather than from the controller. Finally, $\texttt{image\_horizon} = 2$
means the Memory axis measures the model together with this context policy. The scores
cannot separate these interface causes from model failure.
\\[2pt]
\textbf{Comparability.} The LLM controllers receive a wiki-derived task
description and process-score ladder in addition to the official instruction. The
public policy cells were not re-run with equivalent text. The comparison therefore
shares the evaluator and outcome metric, but not the input information. Those cells
are also a leaderboard snapshot and were not re-run alongside the LLM-controller
columns. The 33 real-robot trials retained around a safety stop are a selected sample
supporting no estimate of general real-world reliability. For a closed-weight model,
contamination by benchmark-related pretraining material is unquantified rather than
ruled out.
\\[2pt]
\textbf{Cost and time scale.} The per-episode call budget is a condition, not a
neutral setting. An episode that ends without success contributes zero to SR,
whether it exhausted its budget or failed for another reason; partial-credit
Scores can still differ. Inference latency and action timing may also limit
deployment. This evaluation does not isolate their contribution to failures
from limitations in perception or physical reasoning.
\\[2pt]
\textbf{The ICL probes are incomplete.} The demonstration-conditioned probe is a
single insertion recipe on one set of tasks and one choice of example. A different
place in the message sequence, a different difficulty slice, or a different
demonstration could change the result. This evaluation therefore cannot conclude
that few-shot ICL does not work. It shows only that one demonstration did not help
under this protocol. The perturbation probe has no history ablation and cannot
separate learned adaptation from existing robustness or ordinary feedback control.
The negated-coordinate condition also exposes transformed tool bounds.
\\[2pt]
\textbf{The main conclusions come from one simulator.} The ranking and the
capability split come from RoboDojo-Sim. No comparably broad manipulation benchmark
was evaluated in another simulator. The RoboDojo-Real campaign was
diagnostic and stopped for safety, so the official real protocol was not completed.
The mobile-humanoid recordings and RoboPianist scores use separate settings and
do not enter the official board. This evaluation therefore does not show that the ranking or the capability
split transfers to another simulator or to hardware under the official real
protocol. Other simulators and real-robot experiments need further investigation.

\section{Ethics statement}
\label{sec:app-statements}

Real-robot evaluation was stopped for safety. During those trials the model
repeatedly issued physically unreasonable or unsafe actions, and some incidents
damaged equipment; no person was injured. These incidents show that the tested
system was not sufficiently safe for continued hardware evaluation. They motivate
independent safety checks and stopping mechanisms rather than reliance on the
model's judgment alone. Any safety intervention should be documented when
interpreting controller performance. The retained 33 diagnostic trials are a
selected sample, not a completed benchmark evaluation
(Section~\ref{sec:real}, Appendix~\ref{sec:limitations}).

The study evaluates commercial models on a public manipulation benchmark and
involves no human subjects and no personal data.


\end{document}